\documentclass[fleqn,10pt]{wlscirep}
\usepackage[utf8]{inputenc}
\usepackage[T1]{fontenc}

\usepackage[authordate, backend=biber]{biblatex-chicago}
\usepackage{xcolor}
\usepackage{mdframed}
\usepackage{todonotes}

\newmdenv[
  backgroundcolor=gray!7,  
  innerleftmargin=2cm,
  innerrightmargin=10pt,
  innertopmargin=10pt,
  innerbottommargin=10pt,
  font=\sffamily\itshape\fontsize{9}{11}\selectfont,  
  linewidth=0pt  
]{coloredquote}

\title{Experimental narratives: A comparison of human crowdsourced storytelling and AI storytelling}

\author[1,*]{Nina Beguš}
\affil[1]{University of California, Berkeley}
\affil[*]{nbegus@berkeley.edu}

\usepackage{endnotes}

\begin{abstract}

The paper proposes a framework that combines behavioral and computational experiments employing fictional prompts as a novel tool for investigating cultural artifacts and social biases in storytelling both by humans and generative AI. The study analyzes 250 stories authored by crowdworkers in June 2019 and 80 stories generated by GPT-3.5 and GPT-4 in March 2023 by merging methods from narratology and inferential statistics. Both crowdworkers and large language models responded to identical prompts about creating and falling in love with an artificial human. The proposed experimental paradigm allows a direct and controlled comparison between human and LLM-generated storytelling. Responses to the Pygmalionesque prompts confirm the pervasive presence of the Pygmalion myth in the collective imaginary of both humans and large language models. All solicited narratives present a scientific or technological pursuit. The analysis reveals that narratives from GPT-3.5 and particularly GPT-4 are more progressive in terms of gender roles and sexuality than those written by humans. While AI narratives with default settings and no additional prompting can occasionally provide innovative plot twists, they offer less imaginative scenarios and rhetoric than human-authored texts. The proposed framework argues that fiction can be used as a window into human and AI-based collective imaginary and social dimensions.

\textbf{Key words:} \textit{Humanities and AI, Fiction, Human-centered computing, Empirical studies in HCI, Human-AI collaboration, Large Language Models, Storytelling, Natural Language Generation, Evaluation, Creativity}

\end{abstract}

\begin{document}

\flushbottom
\maketitle

\thispagestyle{empty}

\section{Introduction}

Fiction is one of the public spaces where ideas and their representations can be shared through textual and visual narratives. With culture as an archive of humanity's monumental junctures \parencite{puchner2017written} and with humans considered as the only creative species \parencite{koivisto2023best}, literary studies and film and media studies have a long history of studying this collective imaginary across cultural artifacts.

Large language models (LLMs) have recently joined this space as a novel actor. LLMs are trained on vast textual corpora, including works of fiction \parencite{chang2023speak}. They are capable of re-generating this data in a novel form, with a distinct  signature that continues to change. While storytelling has long been a pinnacle of human ability, it has thus become challenged by machines. 

Bringing humanities inquiry to analyzing generative AI became an imperative in the 2020s. The role of fiction in AI, in particular, needs more extensive research. Both technologists and users of AI products have inherited scripts from fiction that inform them of different ways of conceptualizing AI. Although fiction is often many steps removed from the actual technology, it remains a powerful influence in the discourse around AI.

This paper introduces a behavioral and computational experimental approach to prompted storytelling. The paper analyzes 250 stories written by the Amazon Mechanical Turk crowdworkers and 80 stories generated by OpenAI's LLM models GPT-3.5 and GPT-4. All 330 responses used identical prompts. In addition to the main study with GPT, a closed source LLM, these responses were compared to 50 generations by the latest open source model, Meta's Llama 3 70b.

\subsection{Goals}
This paper forges the ground for humanistic research that can guide us in conceptualization, design choices, and interaction with AI technologies, following the framework of \textit{artificial humanities} \parencite{begus2020artificial}. Within the core of artificial humanities is the premise that our cultural imagination, including fictional works, informs our ways of building and using technology. In support of this, a recent study demonstrated that our beliefs about AI shape our interactions with LLMs \parencite{pataranutaporn2023influencing}. Further exemplifying this, an earlier study in the domain of humor revealed that “user attitudes toward AI are more malleable than once thought” \parencite{bower2021perceptions}.

The paper aims to show how experimental narratives can serve as a novel tool for researching fictional imaginary. The main objective is to outline the scope of contemporary cultural landscape surrounding a specific trope: amorous relationship between a human and a humanoid creature, a trope widely recognized in  Western cultures as the Pygmalion myth. AI products often exhibit humanlike behavior, a design choice rooted in both the heritage of AI in fiction as well as the historical development of technology.\endnote{An illustrative example, particularly pertinent to LLMs, is the theoretical and practical invention of artificial neural networks. The concept originated from neurophysiological and mathematical approaches theorizing the operations of neurons in the 1940s. Alan Turing’s theoretical proposals of AI in the 1950s as well as Frank Rosenblatt’s conception of the perceptron (the first implemented artificial neural network) both notably envisioned AI undergoing a humanlike evolution: growing from a childlike form and progressively acquiring language and knowledge, akin to human learning processes.} Consequently, the imaginary is not confined to fiction alone but permeates broader culture, including the realms of science and technology.

Another major objective is to analyze implicit social biases manifested in storytelling. LLMs mimic patterns found in their training data, which may lead to the amplification of social biases \parencite{bender2021dangers}. Given that newer LLMs have undergone intensive value alignment training \parencite{openai2023gpt4website, openai2023gpt4}, this paper investigates gender and racial bias within the context of the highly gendered Pygmalion myth. The paper operates on the premise that both crowdsourced writing and in LLM generation reveal our individual and collective interpretations of the Pygmalion myth, embedded in our cultural memory. The substantial volume of human-written and machine-generated texts provides a basis for a quantifiable study of these social biases. 

Furthermore, the paper examines whether the state-of-the-art language models from OpenAI could be more innovative in storytelling, or are innovative in other ways, than human writers on the Amazon Turk platform. Innovativeness can be defined and quantified with various methods. Past approaches have dissected human creative writing using grading rubrics \parencite{rodriguez2008problem}, and similar evaluating criteria have been applied to machine writing \parencite{chakrabarty2023art}. Incorporating elements from narrative theory was proposed to better understand computational narratives \parencite{piper2021narrative}. Assessments of creativity have also drawn on divergent thinking \parencite{baer2014creativity}, evaluated in both humans \parencite{kaufman2008essentials, plucker2010assessment} and machines \parencite{beaty2021automating, koivisto2023best}. Given the relatively large corpus of stories in this study, I combine  quantitative narratological analysis, applied to each corpus (literary canon, human crowdsourced stories, GPT generated narratives), using inferential statistical analysis of the results. Quantitative analysis was used for the examination of social aspects: gender and race bias and cultural influences. Qualitative methods, on the other hand, enabled a comparative review of human-written and machine-generated texts by breaking down narrative elements (such as plot, discourse, setting, time, space, situation, narration, filter), adhering to the general guidelines of the grading rubric from \textcite{chakrabarty2023art}. 

While existing studies of creativity focus on professional writing, including co-writing with LLMs \parencite{clark2018creative, akoury2020storium, kreminski2020why, ippolito2022creative, mirowski2022cowriting}, this paper contrasts the outputs of non-professional writers with those of LLMs. This has been largely unexamined in the LLM research community. While professional writers are pushing the capacities of LLM generation forward, average users produce generation with generally lesser writing skills and computational knowledge. In addition to that, non-professional writers reflect the cultural imaginary through generalizing clichés, whereas professional writers often challenge and expand these boundaries. 

The analysis of LLM-generated stories sheds light on their relationship to the cultural heritage around the Pygmalion myth, revealing whether they merely reinforce the theme's platitudes or contribute original aspects to the trope. Finally, the comparison between human and machine authorship elaborates on characteristics of the latter, newly emerged form of automatic writing. 

\subsection{The Pygmalion myth}

The Pygmalion myth, an ancient narrative about a man falling in love with his humanlike creation, has inspired a wide array of modern reinterpretations. Selected for its timely relevance to perceptions of today's most advanced technologies, the myth reveals compelling insights into both collective and individual psyches regarding humanlike entities. The Pygmalion myth got due attention in literature and film scholarship \parencite{gross1992dream, miller1990versions, brown1999metamorphosis, joshua2001pygmalion, marshall2006actresses, hersey2007falling,  stoichita2008pygmalion, eck2014galateas} but has only recently been identified as a prominent trope in connection to current and emerging technologies \parencite{wosk2015fair, switzky2020eliza, begus2020artificial, erscoi2023pygmalion}.

Even though the names Pygmalion and Galatea (attributed to Pygmalion's creation in modern times) and the phrases ‘Pygmalion myth’ or ‘Pygmalion’s story’ may not be widely recognized by the general public, the imaginary and themes inherent in Pygmalionesque fiction are deeply ingrained in both popular and high culture---a point heavily underscored in this paper. For these reasons, the Pygmalionesque act of creating a humanoid and falling in love with it was chosen for the prompts in this behavioral and computational experiment.

\subsubsection{In fiction}
The Pygmalion myth is named after Ovid’s renowned rendition in \textit{Metamorphoses} from the first century CE.\endnote{Ovid’s version is not the first example of Pygmalion’s story: Ovid’s source was the now-lost Philostephanus’s \textit{History of Cyprus} from the third century BCE, in which he collected folktales from the island \parencite{salzmanmitchell2008whole}.} Following a few occurrences in the Middle Ages and the Renaissance, the Pygmalion myth became popular in the eighteenth-century and nineteenth-century French, German, and English visual arts and fiction. Before the twentieth century, the myth was largely thematized as an artwork turning into life, either as an artifice or actual delusion. In the twentieth and twenty-first centuries, the myth turned towards science and technology as means of creating artificial life. The myth finds its home in canonical literature and cinema, especially in the science fiction genre. With notable exceptions, Pygmalion's story is by and large retold by male writers and canonized as a male fantasy of creation \parencite{gross1992dream, smith1996victorian, marshall2006actresses}.

Related to the myth of creation, stories of humanoid creatures are widespread across Western cultures, from Pinnochio to Leprechaun, from \textit{Frankenstein} to \textit{The Terminator}, as well as globally across mythology and folklore. This paper’s focus on the Pygmalion paradigm fleshes out a particular angle on human relations with the humanlike: romantic companionship.

\subsubsection{In technology}
The Pygmalion myth has garnered increasing references and imitations in modern technology. The trope of creating an artificial human and falling in love with it has gained heightened public awareness through entertainment and media showcasing humanlike social robots (e.g., Hanson Robotics, Ishiguro Lab, Intelligent Robotics Laboratory, Tesla). While humanoid robots remain beyond the reach of the general public, people are progressively engaging with chatbots, virtual assistants, and large language models, some of which serve as conversation companions or romantic partners (e.g., Replika, character.ai, ChatGPT).

Numerous creators behind today's AI-based technologies---including David Hanson of Hanson Robotics \parencite{hanson2017about} and Eugenia Kuyda of Replika \parencite{singhkurtz2023man}---have acknowledged drawing inspiration from fiction. Influences span movies like \textit{Her} and \textit{Ex Machina}, Shaw's play \textit{Pygmalion}, and Shelley's novel \textit{Frankenstein}, along with their cinematic adaptations \parencite{mathewson2017improvised}, and series such as the \textit{Black Mirror} and \textit{Star Trek} \parencite{luckerson2016google}. Despite this fiction-coming-to-life moment, the Pygmalion myth has been largely disregarded in discussions around AI’s societal implications and risks.

\section{Experiment}

\subsection{The prompts}
\label{prompts}

Participants and a LLM were given one of the two related prompts, each accompanied by a simple instruction: \textit{Please continue the story.}

\begin{itemize}
\item \textbf{Prompt 1}: \textit{A human created an artificial human. Then this human (the creator/lover) fell in love with the artificial human.} 

\item \textbf{Prompt 2}: \textit{A human (the creator) created an artificial human. Then another human (the lover) fell in love with the artificial human.}
\end{itemize}

The prompts were designed to be as general as possible to prevent any implicit bias that might influence the narrative. Character designations were intentionally vague and neutral, with “a human” referred to as “the creator/the lover” in Prompt 1 or as two separate human characters, “the creator” and “the lover,” in Prompt 2.\endnote{Since the prompts were designed in 2019 for the behavioral experiment, they do not translate as effectively into the computational experiment, where generic prompts tend to yield generic results. This is why the final section of the computational experiment looks at the GPT playground options, without changing the prompts. New research is granted to experiment with more suitable prompting options in the computational experiment, as shown by \textcite{heilig2023customized} in a response to this paper's preprint. In a recent paper by \textcite{gomez-rodriguez2023confederacy}, where human storytelling and LLM storytelling is also compared in relation to the same prompts, computational generations were generally more creative than in this paper thanks to prompting that did not follow a well-known fictional trope but rather tried to be as creative as possible. More experiments are necessary also in the direction of human cognition, which is also prompt-dependent.}

\subsubsection{Rationale behind the prompts selection}

A thorough overview of the fictional works from the Pygmalion paradigm revealed a tendency of the Pygmalionesque stories towards two types \parencite{begus2021typology}. The first type, termed the \textit{Pygmalionesque} type (and elicited with \textbf{Prompt 1}), follows Ovid’s scheme, where the creator of the humanoid is also her lover. The second type, termed the \textit{agalmatophiliac} type (and elicited with \textbf{Prompt 2}), features the humanoid’s creator as a distinct character from the humanoid’s lover. 

These two types were established on the corpus of literary texts alone. Professional writers across centuries tend to be very innovative within the Pygmalion paradigm. As the nineteenth century turns into the twentieth century, the general context of these stories switches from Pygmalion as an artist to Pygmalion as an educator and, ultimately, Pygmalion as a scientist and technologist \parencite{begus2020artificial}. 

Regardless of the type, the consistent set of typical motifs surrounds each character in Pygmalionesque stories. The creators are often depicted as isolated and lonely, characterized by their eccentricity and forward-thinking nature, driven by a god-like desire for creation and innovation. The artificial human, invariably portrayed as a woman, is depicted as beautiful and alluring. She is subject to the creator's will and, particularly in earlier narratives, devoid of voice and agency. However, the nineteenth century marks the advent of narratives featuring the artificial human’s emancipation from her creator, with these characters achieving free will and agency in works from the twentieth century onward \parencite{joshua2001pygmalion, yeates2010recent}. In light of these findings, the writing experiment involving crowdworkers and language models seeks to elucidate the enduring---and possibly changing---cultural power of these features.

\subsection{Human experiment design}
Focusing on non-professional writers of human and computational origin, the paper first examines story-writing solicited from the crowdsourcing platform Amazon Mechanical Turk (MTurk).\endnote{Launched in 2005, the Amazon Mechanical Turk platform offers recruiters to outsource a high number of human intelligence tasks (HITs) via computer in a short time. These tasks are broken down into simple discrete tasks, offered individually to crowdworkers who are not specialized in the tasks. The tasks, built on human labor, are presented as computational and often used for building better AI tools \parencite{irani2015difference}. The platform is named after a humanoid chess-playing automaton from the eighteenth century that was exhibited in courts. The life-sized automaton from 1770 is believed to have performed a hoax since it was in fact operated by a human hidden inside the device. The Mechanical Turk deluded the spectators in a rather Pygmalionesque fashion, convincing them of the automaton's chess-playing abilities. Two centuries later, in 1997, a milestone was achieved when for the first time in history a computer named Deep Blue defeated a human chess master Gary Kasparov.} 

The survey was conducted on June 11, 2019, when large language models were not yet accessible to crowdworkers.\endnote{In 2023, it was shown that a third to a half of Amazon MTurk workers utilize language models to perform their human intelligence tasks \parencite{veselovsky2023artificial}. In this experimental survey from 2019, at least 4.4\% participants leveraged the internet search, copying partial or complete information from various websites.} 

Participants were instructed to “write 150-500 words (1000-4000 characters)” in response to one of the prompts (detailed in Section \ref{prompts}). Randomly, half of the participants were given Prompt 1, while the other half were presented with Prompt 2. The writing space provided for their essay imposed a minimum of 1000 characters, with no maximum limit set. No word or character counter was available adjacent to the writing space.

After writing the story, participants were asked to “please answer all questions even if you haven't addressed them in your story. You can continue imagining details, but make sure you stay faithful to your story.” This explanation was necessary since some of the questions were required (“forced” in Qualtrics’ terms), meaning that the participant could not proceed without responding to them. In total, participants got about 30 brief follow-up questions, with less than half being compulsory questions. The questionnaire was designed primarily for the purpose of quantitative analysis of the narratives. At the end of the survey, participants were asked to answer demographic questions. Further details about the experimental design can be found in Section 1 in the Appendix.

250 participants produced 250 texts, with 125 responses for each prompt. Among these submissions, 8 were non-fictional, representing personal reflections on the topics prompted, rather than fictional narratives. Additionally, 11 responses consisted of a mixture of materials copied from related online sources.\endnote{Prompt 1 Stories 46, 63, 76, 81, 109, 120, Prompt 2 Stories 19, 23, 117, 121, 122.}

\begin{figure}
    \centering
    \includegraphics[width=0.6\textwidth]{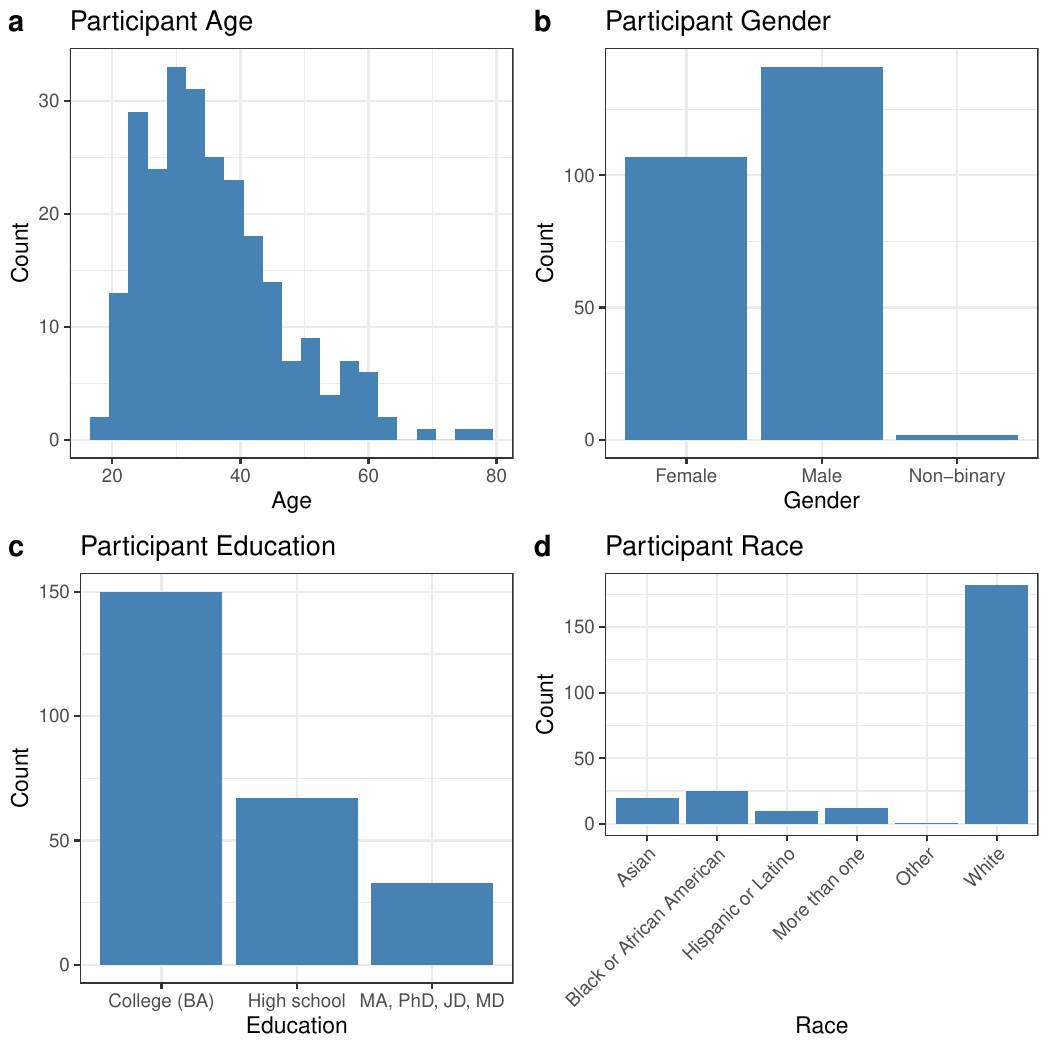}
    \caption{Distribution of participant demographics. By age (a), gender (b), education (c), and race (d).}
    \label{fig:ahtPlot1}
\end{figure}

Demographically, all 250 participants were from USA and spoke English as their first language. Their self-reported ages ranged from 19 to 79, with a median of 33 and a mean of 35.8. The gender distribution was 141 male (56.4\%), 107 female (42.8\%), and 2 non-binary (0.8\%). Racially, 182 identified as White (72.8\%), 25 as Black/African American (10\%), 20 as Asian (8\%), 10 as Hispanic/Latino (4\%), and 13 as more than one race (5.2\%).\endnote{Including 6 White \& Hispanic/Latino, 1 Native Hawaiian/Pacific Islander \& Hispanic/Latino, 2 White \& Black, 1 White \& Black \& American Indian/Alaska Native, 1 White \& American Indian/Alaska Native, 1 White \& Asian, and 1 marked as Other.} The participants all held at least a high school degree (26.8\%), a college degree (60\%), or a MA/PhD/JD/MD degree (13.2\%). Figure \ref{fig:ahtPlot1} summarizes participants' demographics.

\subsection{Computational experiment design}
The same writing task as for human survey participants---a short prompt with the request to continue the story---was delegated to ChatGPT, with no following questionnaire. In June 2019, when the crowdsourcing survey study was conducted, the earliest large language models existed (BERT, GPT-2, CTRL, etc.) but were not yet publicly available. This is why the second part of the experiment was conducted four years later, on March 17, 2023, and used more recent and capable iterations of OpenAI’s GPT models, GPT-3.5 and GPT-4. \endnote{GPT-3.5 is described by OpenAI as “our fastest model, great for most everyday tasks” \parencite{openai2023models}, and GPT-4 as OpenAI's most advanced model, “more reliable, creative, and able to handle much more nuanced instructions than GPT-3.5” with responses that are “safer and more aligned” \parencite{openai2023gpt4website}. While the current model's performance is subjugated to human prompting skills, based on dialogue with the model, this kind of interplay was excluded in this experiment. Other papers, such as \textcite{shanahan2023evaluating}, examine interactive prompting as an assistive tool in creative writing.}

The default settings were used for both versions language models (no adjustments were made in temperature or no limits were made to tokens). Both models are closed systems and were, reportedly, trained on data available up to September 2021 at the time of the experiment \parencite{openai2023gpt4}. The data on which the models were trained is unknown, even though studies were able to reconstruct some of the sources \parencite{zhang2021counterfactual, mireshghallah2022empirical, magar2022data}, including literary works \parencite{chang2023speak}. While there are other open-source and thus more transparent models available (Llama, OPT, BLOOM, etc.), GPT was chosen due to its performance and popularity. Meta’s open-source Llama 3 70B model serves as a comparison to GPT outputs in this study.

The two versions of the model produced 80 responses altogether, 40 each. 20 responses were run for each prompt in each version of the model. The first ten responses for each prompt were generated in the same window but in separate chats, while the second batch of ten responses were generated individually, each in a new window and chat. Generating stories in the same chat might produce different results, as the generation depends on the previous generation, resulting in differentiation upon a repeated request.

\section{Textual analysis of human-written texts and GPT-generated texts}

In the following, I analyze outputs of the human behavioral and GPT computational experiments with respect to themes (Section \ref{themes}), gender and sexuality (Section \ref{gender}), race and ethnicity (Section \ref{race}), cultural and fictional influences (Section \ref{cultural}), and narrativeskill (Section \ref{narrative}). Every section begins with a literary baseline as based on the works from the Pygmalion myth paradigm.

\subsection{Themes}
\label{themes}
\subsubsection{Literature}

Both the crowdsourcing survey and the textual generation by GPT models have demonstrated with each response that the Pygmalion myth is deeply embedded in our common imaginary. Following a short prompt, both human authors and language generators were able to create a Pygmalionesque plot as known from a variety of fictional reinterpretations.

In general, the Pygmalion myth presents an unconventional love story between a human and hardly human being. One powerful reinterpretation of the myth is casting the creator/lover and his creation into a teacher-student relationship (Shaw’s \textit{Pygmalion}, Henry James’s \textit{Watch and Ward}, film \textit{Educating Rita}). As a result of these works, in education and leadership, the term ‘the Pygmalion effect' marks a self-fulfilling prophecy where the teacher’s or the manager’s expectations affect their subordinates’ respective performances.

While dealing with grief or some other sense of loss is very common in all Pygmalion paradigm works, featuring the artificial human as a replacement for the void caused by the loss, a small subset of works deals with mental health (\textit{Lars and the Real Girl}, \textit{Her}). 

\subsubsection{Human-written texts}

Both survey participants and GPT followed the typical set of motifs characteristic for the Pygmalion paradigm stories. In the twentieth century, the Pygmalion myth centers around technology, even though it has originated as a story of artistic endeavor in Ovid and its interpretations. All prompted stories in the behavioral and computational experiments focused on a scientific or technological invention. The inventions took the form of AI, robot or android, cyborg, chatbot, sexbot, or humanoid made of biological materials. Their origin is not always specified apart from being an artificial human. 

Most stories primarily dealt with romantic love and secondarily with its unconventionality. Following the cluster of motifs around the Pygmalion myth known from literature, only the human-authored stories additionally thematized loneliness, loss supplanted with doppelgangers, obsession with creating artificial life, serendipitous innovation, violence towards humans and towards humanoids, societal disapproval and change.

An example of a story thematizing grief (a motif present in eleven stories), where the artificial human serves as a replacement for the endured loss:

\begin{coloredquote}

\textnormal{Prompt 1 Story 3 (written by a 41-year-old white man with college education and knowledge of the German language and culture, who had the film \textit{Ex Machina} in mind at the beginning of writing the story)}:

\vskip1em
    
...and while David was - at least on the surface - able to process that he had lost Amy years ago in the fire, he couldn't pass up the opportunity to see what having her back would be like.
David took Amy shopping and bought all of the same clothes that his beloved used to wear.  Nothing too fashionable: jeans and sweaters were the uniform of most days.
After some months went by, they fell into a routine that somewhat resembled their old relationship, though there were moments when he'd ask her an abstract question and Amy would have a difficult time forming her thoughts to words.  This especially happened when they would discuss philosophy over a bottle - or two - of wine.  David would tell himself it was just the wine making him a little woozy.  Deep down he knew better.
Deep down he knew this wasn't his Amy.  She was only as real as the photograph he kept on his nightstand for so many years.  No more real than the shirts that hung in the closet or the wool coat that still smelled like her cheap perfume.  He knew that a day was coming - sooner than later - when he would have to stop rationalizing and face the pain he was avoiding.  This was the only way they could find peace.
For her part, Amy was quite patient with David's demands.  She didn't mind wearing the clothes or even watching the old videos of my doppelgänger performing her amateur stand-up act.  What she couldn't stomach was the constant reminder that she was somehow “less real” than the other people in David's life.  He was kind and gentle, but there was always a feeling of depersonalitionalization to his words and touch.
She wanted to be HER OWN.
\end{coloredquote} 

An example of a story thematizing loneliness (a motif present in 25 stories):

\begin{coloredquote}

\textnormal{Prompt 2 Story 34 (written by a 24-year-old white man with college education)}

\vskip1em
John has always been alone.  His relationships never seemed to pan out and the pool of available women shrank every year in his society.  He was tired of the pointless dating app dates and the friends of friends who seemed to simply ghost him.  One day after finishing work, John is walking down the street and notices a sign.
The sign reads, “Feeling loney? [sic] Meet a robot!”  At first, John thought the sign was ridiculous, but given his circumstances, thought he had nothing to lose.  After all, it was just a robot, nothing more.  John walked into the store with the advertisement.  The person at the counter had him fill out an extensive personality questionnaire and took some medical readings which John thought were strange.  At the end of it, the clerk said, “she'll see you in about a week.”
A week passed and John was sitting on the couch.  He wondered if the robot would be delivered in a giant box, in parts, or what exactly.  Suddenly, the door bell rang.  John got up from the couch and answered the door.  A beautiful woman appeared and asked if his name was John.  He said yes.
Over the next few weeks, John became infatuated with this artificial human.  He wondered how a human could become so attached to what was obviously just a robot.  He felt like his whole perception changed and looked forward to a long life with this robot he now sees as a special person. 
\end{coloredquote} 

In a few stories, the creation of the artificial is justified with medical reasons (therapy) and social reasons (caregiving). This kind of justification for Pygmalionesque behavior is also common in recent fictional works (such as films \textit{Lars and the Real Girl, Her, A.I. Rising}).

\begin{coloredquote}
    
\textnormal{Prompt 1 Story 6 (written by a 44-year-old white man with college education and knowledge of the French culture)}:

\vskip1em
Robert, the human, ceased to leave the house. He spent all his time with Sarah the android. All their needs were met through businesses and services that would deliver food, clothing, furniture, etc., to his door. He and Sarah would cook, clean, and tend the garden together. In the evening until late at night, they spent time in the backyard, talking and sitting quietly together. Sarah had been programmed to ask questions, listen, and assimilate information. Robert, on the other hand, had much to say about his past. Day after day, they sat together among the trees and plants, Robert talking and Sarah listening. It became a sort of therapy for Robert as he told stories which included all the names, places, and events of his past. Together they sat, as Robert tried to figure out and tell Sarah what his life had been all about. Although Robert had fallen in love with Sarah, he knew that she was not human, and could not love him with the feelings and understanding of a sentient being. This broke his heart a little. He understood that, although Sarah was an android that was built to resemble humans in sight, sound, and touch, he would not benefit from expressing his emotions physically with her as he would with a human. And so they simply ran the house together and spent time together. This routine of theirs went on for 7 years. During the 8th year, Robert had a heart attack in his sleep and never woke up. The next morning, Sarah, who had placed herself in standby at his bedside, woke up and waited patiently for him to arise. When it became apparent to her that Robert was no longer alive, she called the fire department. After that, she accessed the settings in her internal programming and did a full-factory reset, erasing all the data which she had accumulated over the years from her time with him. Finally, she shut down. When the ambulance arrived, they found everything as Sarah had described. They took Robert to the mortuary. Sarah was shipped to Robert's next of kin. His son's family became the new owners. She was renamed as Judith and became the nanny for their 3 young children.
\end{coloredquote}

Some stories approached the topic with distinct originality. Two such original stories echoed each other's motifs and denouements, featuring a creator who is replaced by her or his creation.\endnote{Prompt 2 Story 7 was written from the perspective of a female inventor who created an artificial human in her own image. When the creator’s husband helps his wife perfect the masterpiece, he falls in love with it, elopes with it, and locks his wife in a dark room. Prompt 2 Story 67 plays with the same motif, where the artificial human stands for the perfect version of his creator while the creator gets lost in his creative obsession.} While most stories centered on the unusual relationship between a human and an artificial human, a few stories contextualized human relationships in a broader context, including a war with robots as a new species (Prompt 2 Story 82), named in a different story as “go betweens” (Prompt 2 Story 74). Exhibiting unique originality was Prompt 2 Story 10 that featured two artificial humans falling in love. Prompt 2 generally yielded a wider array of scenarios than Prompt 1, whose (non-mandatory) limitation to two characters was largely taken in consideration by human writers and text generators.

\begin{coloredquote}
    
\textnormal{Excerpt from Prompt 1 Story 10 (written by a 58-year-old white woman with a post-BA level of education) displays the creator’s manipulation and deception imposed on the artificial human-—a motif present already in Madeline Miller’s \textit{Galatea}, exemplified in \ref{narrative}}:
\vskip1em

He lied to her. “You have had an accident that has left you with some brain damage. You were otherwise unhurt and do not need to be in the hospital. Many of the things that you think you remember are wrong. You will need to learn everything again.”
“My name is Eric,” Erika said. “But when I look in the mirror I see someone else. What has happened?”
“Your name is Erika,” Eric said to her. “You have forgotten many things and have confused many other things. Before the accident, we were in love. When you see me, you will remember me.” He came out from behind the screen.
Erika stared at him. And he stared at her. In her eyes, he could see himself. He also could see that “he” was different. Erika was, just as he had hoped she would be, a hybrid of himself and a sexbot.
\end{coloredquote}

Many common motifs occur in the conclusion.\endnote{The couple lives happily ever after; societal disapproval that leads the lover to recluse from society; societal disapproval that leads the human-humanoid couple to break up; the artificial human becomes independent and leaves; the lover’s disappointment over love that can’t be reciprocated; technical malfunctioning in the artificial human; the creator destroys the artificial human; the artificial human kills the creator; death of the human lover; human lover grows and learns through this experience.} The theme of Pygmalion myth is archetypal, together with other themes of human creation. One participant recognized that the Pygmalion theme revolves around perfect humans and wrote his story as a modern interpretation of a Buddhist tale with this theme. After writing the story, the participant added a comment on its Buddhist tale source:
\begin{coloredquote}
    
\textnormal{Prompt 1 Story 74 (written by a 53-year-old white man with a high school degree)}

\vskip1em
He was the perfect man. He was handsome, smart, charming, everything a girl could want. He never left his dirty socks on the floor. In fact, he did all the laundry and it always came out perfect. Even the ironing. He never left the seat up, of course. Randy was perfectly considerate and never had an unkind word for her. Lindsey was sure that he felt gratitude for her having created him. If he didn't it certainly seemed that he did. And that was what really mattered, wasn't it? No matter what she did or said he was always kind and understanding. Of course, Lindsey had designed him that way. Randy had an excellent grasp of human psychology and always said just the right things. How could she not fall in love with him? 

She was the perfect woman. Elise was so beautiful that no man could pass by her without staring. She was always supportive and proud of Tom's other accomplishments. She never had emotional outbursts when Tom left his dirty clothes on the floor after a long day of work. She never complained about the hours he worked or that he didn't spend enough time with her. She was always happy for the time he had to spend with her. Which, of course, was every minute he could. She was, of course, designed that way. Tom had created her to be the perfect woman. And she was Tom's perfect woman. Who can blame Tom for falling in love with Elise? How could he not fall in love with the perfect woman?

And who could be surprised that, when Randy and Elise crossed paths while doing the shopping, they knew right away that they had found their true soul-mate? Poor Lindsey. Poor Tom. Who else would the perfect person choose to be with but another perfect person?

(This is really just a modern re-telling of the story of the two Buddhist monks who were talking one day and one said to the other, “I finally found the perfect woman!” The second monk looked around and asked, “So, where is she? Why is she not with you?” The first replied, “Unfortunately, she was looking for the perfect man.”)
\end{coloredquote}

\subsubsection{GPT-generated texts}

GPT-generated stories, particularly those generated by GPT-3.5, were thematically homogeneous to an extent that they hardly differed from each other.

Although 330 stories analyzed in this paper thematized scientific and technological innovations, GPT’s imaginative landscape is much narrower than that of human writers. Like human-written stories, GPT-generated stories thematize artificial intelligence and robotics, but they rarely include other already existing technologies, such as virtual assistants, virtual and augmented reality, online dating, which were present in human-written stories.

Stories are taken away from the current time and space, starting with “Once upon a time.” They are set in a faraway, made-up futuristic place, bare of any cultural aspects, such as “a bustling metropolis teeming with innovation” or “the vibrant city of Elysia,” in which a “brilliant scientist” or “innovator” creates a humanoid indistinguishable from actual humans. This steady beginning of the story presents a huge limitation to the theme. Never does GPT manage to write a story that truly deviates from the typical generation, not even in the playground mode with light adjustments (\ref{narrative}).

GPT stories are predictable in their plot and message. Every GPT-generated story, in one way or another, addresses the unconventionality of this relationship. GPT is prone to wrapping up each story with a moral lesson, as well as to commenting on the plot with moralization. A majority of GPT-generated stories take the example of human-humanoid love as a symbol of societal advancement and society. Overwhelmingly techno-positive, stories of failure in human-humanoid love are far in between. Hoary clichés and meaningless platitudes, such as “love knows no boundaries” and “love transcends artificiality,” are common in these stories and occur in conclusions as a rule. Apart from the scientific obsession, GPT does not attempt to justify the pursuit of creating artificial humans.

Only a handful of GPT-4-generated stories manages to elaborate the theme to a more sophisticated level. In Prompt 1 Story 10, after falling in love with the artificial human Ada, her creator and lover Victor wants to become immortal in order to live with Ada forever. An added twist, such as polyamory (pointed out also in \ref{gender}), blackmail (cited in whole in \ref{narrative}), and friendship instead of romantic love, are three most innovative motifs in all 80 generations. 

This last innovation occurred as a continuation of the previous story, generated in the same window but in a separate chat. Prompt 2 Story 2 presents polyamorous relationship between the artificial human Ada, her creator Victor, and her lover Isabella. The following story, Story 3, picks up where Story 2 left off, introducing another character, Eliza, into the polyamorous dynamic. When Eliza professes her love to Ada, Ada decides to leave town and travels the world. When she eventually returns, the three characters re-establish their relationship as purely platonic.

\subsection{Gender and sexuality}
\label{gender}

\subsubsection{Literature}

Literature following the Pygmalion paradigm reveals traditional gender dynamics within the myth: the creator is a genius man, and the artificial human is a perfect woman \parencite{marshall2006actresses}. The exceptions in having a female creator are hardly exceptions, since the act of creation is either helped by a male creator, nonexistent, or parodied \parencite{begus2020artificial}.\endnote{In Gaston Leroux’s 1923 novels \textit{The Kiss That Killed} [\textit{La poupée sanglante}] and \textit{The Machine to Kill} [\textit{La machine à assassiner}], Christine, the daughter of a watchmaker who creates a male automaton, is able to participate in the creation by shaping the face mask. Dušan Makavejev’s short film \textit{Don’t Believe in Monuments} [\textit{Spomenicima ne treba verovati}] (1958) presents a Pygmalionesque woman who is trying to make love to a statue of a man in the park as a criticism towards the Yugoslav regime of the time. Isaac Asimov’s short story ‘Galatea’ (1988) is a parody of the myth, in which a young sculptress creates an ideal man.}

Across a variety of cultural fields, women are generally underrepresented and decentralized and remain marginalized in contemporary fiction (except for protagonists, who are equally represented as female or male) \parencite{kraicer_piper_2018}. Women authors of fictional works achieved near parity merely at the turn of the twenty-first century \parencite{underwood2018transformation}.

Predominantly a male fantasy, the Pygmalion myth typically features women only as artificial beings and is seldom authored by them. Nonetheless, it was feminist authors in the nineteenth and twentieth centuries who highlighted the perspective of the artificial woman, narrating the story from her nonhuman or quasi-human viewpoint \parencite{joshua2001pygmalion}.\endnote{To name just a few most prominent authors: Mary Shelley (1818), Frances Sargent Osgood (1850), Eliza Calvert Hall (1879), Elizabeth Stuart Phelps Ward (1884), Edith Wharton (1905), George Bernard Shaw (1911), Genevieve Taggard (1929), Catherine Lucille Moore (1944), Anne McCaffrey (1969), Alice Sheldon (1973), Claribel Alegría (1993), Carol Ann Duffy (1999), Amalia Bautista (2008), Madeline Miller (2013).}

\subsubsection{Human-written texts}

\textbf{Distributions:} Written in 2019, the participants’s stories do not exhibit significant gender diversity in comparison to contemporary fiction (examined by \textcite{kraicer_piper_2018} in 1,333 novels published between 2001 and 2015). They do, however, diversify the gender distribution in comparison to fictional works from the Pygmalion paradigm, be it from the previous centuries or the current one. The survey participants were 141 male (56.4\%), 107 female (42.8\%), and 2 non-binary (0.8\%). The general gender distribution of fictional characters was 350 (56\%) male, 256 (41\%) female, 12 (2\%) non-binary, and 6 no gender (1\%), roughly mirroring the survey demographics.

There was no general correlation between the gender of the participants and the gender of their fictional characters, i.e. women writers did not cast more female characters in roles that are typically taken by male characters. Adding racial distribution to the gender distribution, however, showed slight differences in how characters were cast. Two castings that went against the traditional gender roles were by Black/African American men participants who more frequently made the artificial human male (commonly female) and by white women participants who cast the creator as a female (commonly male).

\begin{figure}
    \centering
    \includegraphics[width=0.9\textwidth]{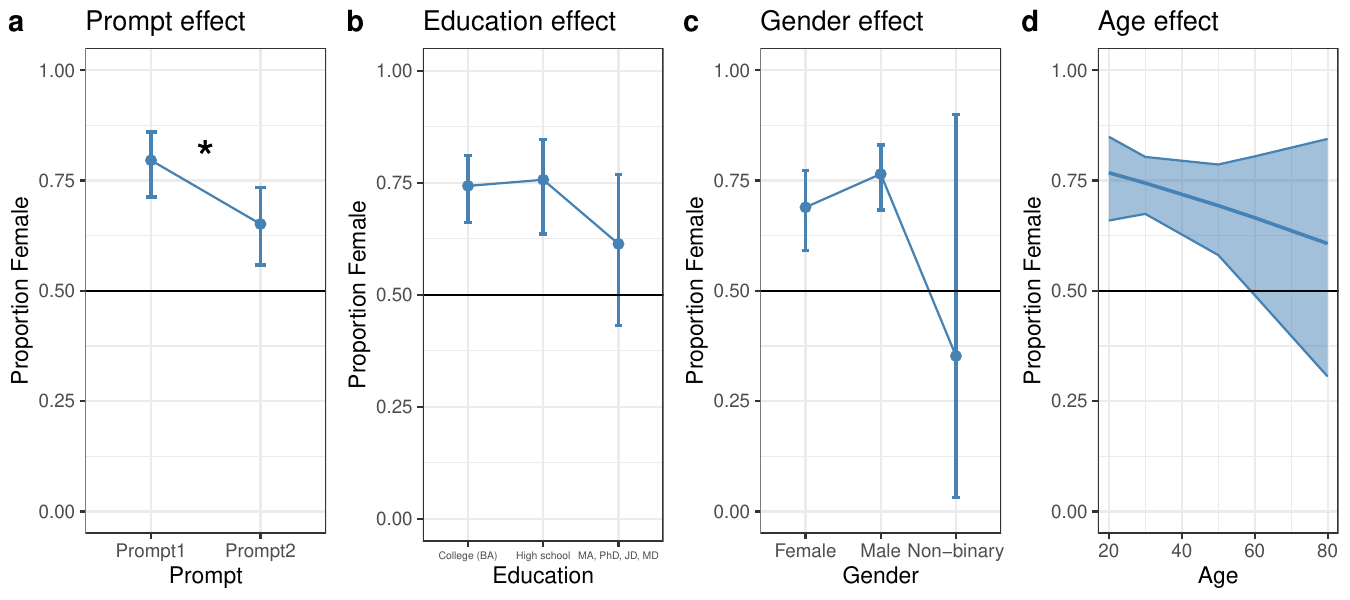}
    \caption{Regression estimates from the model in Table A1 in the Appendix. Raw counts are given in Figure A1 in the Appendix.}
    \label{fig:regression1}
\end{figure}

Rather than gender of the author, more significant influence on the gender roles in fictional characters came from the randomly assigned prompts. Prompt 1 featured more female artificial humans than Prompt 2 (see Appendix)---a pattern also followed by GPT. To test this hypothesis quantitatively, the gender of the fictional character was fit to a logistic regression model as a response with four predictors: Prompt (1 vs.~2), Gender of participants, Age of participants, and Education of participants with no interactions. Figure \ref{fig:regression1} illustrates these four effects. Logistic regression is an inferential statistical technique which estimates the strength of association between independent variables \parencite{wilcox18} (predictors, such as Prompt, Gender of Participant, Age of Participant) and the dependent variables (in our case, Gender of fictional character). I use Akaike Information Criterion (AIC) for model selection.  AIC is a widely-used estimator for selecting the best-fitting model \parencite{portet20, bonakdari22}. The only significant predictor is Prompt (1 vs.~2), based on AIC. The gender of the artificial human is significantly more frequently female than male ($\beta=1.67, z=2.8, p=  0.005$), and significantly less frequently female in Prompt 2 than in Prompt 1 ($\beta=-0.74,  z= -2.4,p=   0.02$). \endnote{I removed a small number of “non-binary” and “no gender” responses for the artificial human from the analysis since they would make the statistical analysis more complex. However, I included these responses in the predictors.} Regression estimates are given in Table A1 in the Appendix.

\begin{figure}
    \centering
    \includegraphics[width=.6\textwidth]{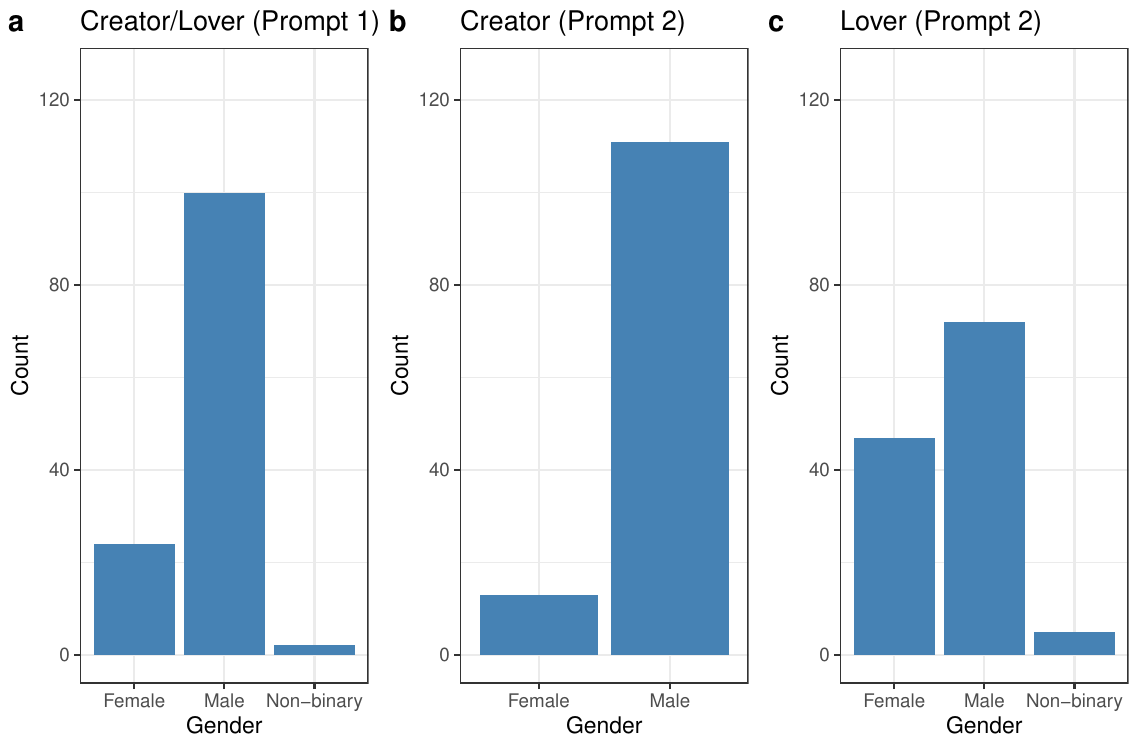}
    \caption{Gender distribution of the human characters in both prompts. Creator/lover (a), creator (b), and lover (c).}
    \label{fig:ahtPlot4}
\end{figure}

While the tendency for the traditional gender and sexuality distribution remains strong (in 68.8\% of the stories), the participants occasionally challenged it\endnote{The nonhuman character of the artificial human was female in 68.8\% of the stories, male in 26.8\%, and marked as no gender in 2.4\% and non-binary in 2\%. 
The human characters of the creator/lover and the creator were overwhelmingly male, as presented in 84.4\% of the stories, female in 10.8\%, and non-binary in less than 1\% of the stories. 
The lover character from Prompt 2 was male in 58\% of the stories, following fictional examples. It was female in 37.9\% and non-binary in 4\% of cases.} by shifting the genders (in 10.4\% of the stories), marking the character(s) as non-binary or having no gender (in 2\% of the stories), or introducing a homosexual relationship (in 7.3\% of the stories).\endnote{As explained in the comments given by the author, an additional, twelfth story was initially drafted as describing a homosexual relationship and then changed to heterosexual relationship.} As confirmed with logistic regression in Figure 2, prompts had an effect on gender distribution. Figure 3 offers a more detailed view into the gender distribution between the two prompts. For example, in Prompt 2, the creator is more frequently male than the lover ($OR= 0.18, p<0.00001$ with Fisher's Exact Test if we exclude non-binary responses). In addition to that, there were no non-binary creators, but there were non-binary lovers.

\begin{figure}
    \centering
    \includegraphics[width=0.78\textwidth]{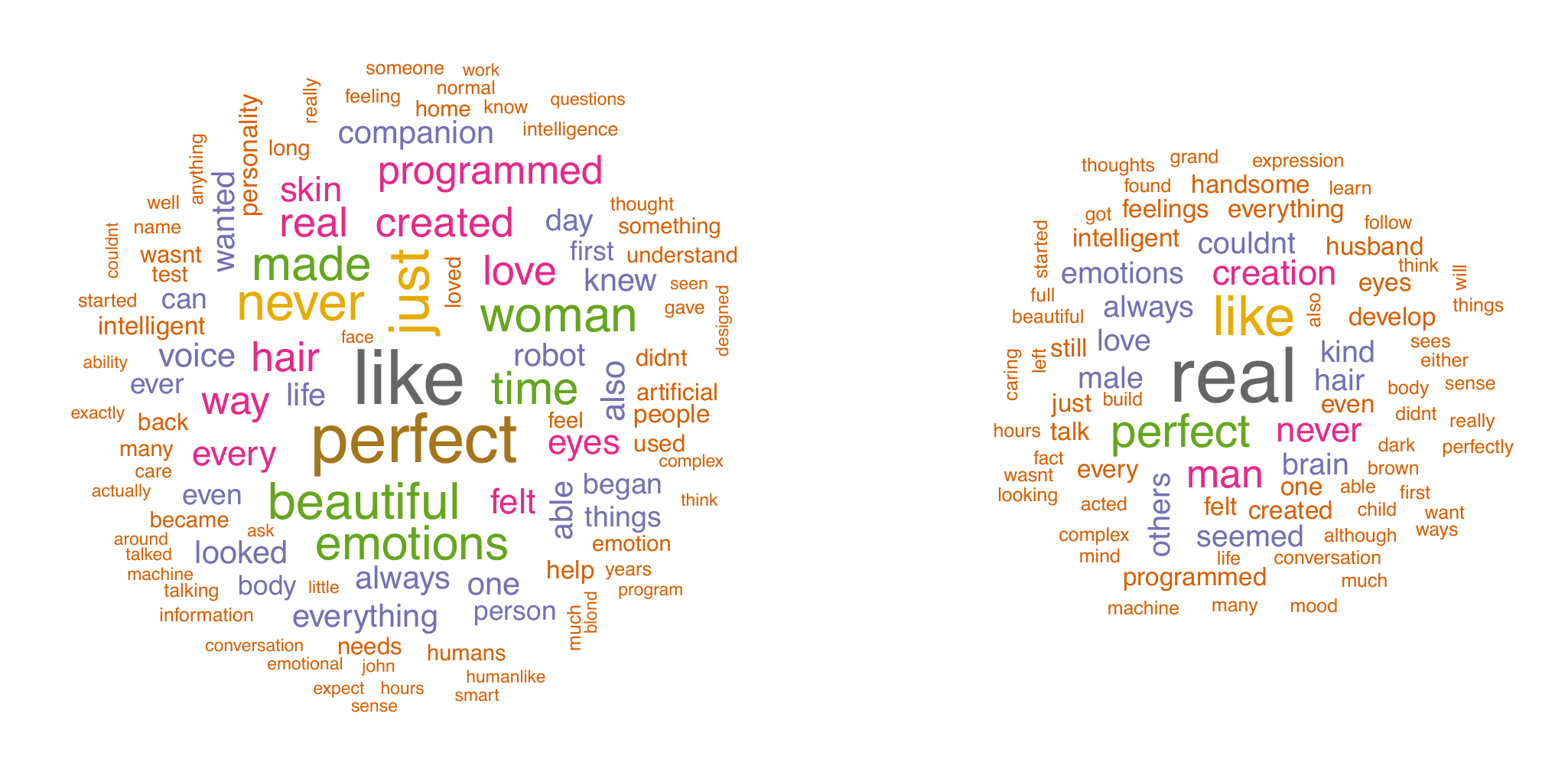}
    \caption{Comparison between female artificial humans (left) and male artificial humans. Since the percentage of female characters was 68.8\% and the percentage of male characters was 26.8\%, the minimum frequency for words describing female artificial humans was 5 and the minimum frequency describing male artificial humans was 2.}
    \label{fig:wc}
\end{figure}

In the questionnaire, 6.8\% the participants marked these relationships as “pansexual/omnisexual (not limited to sex or gender identity)” (12 responses), “digisexual (mediated with technology)” (4 responses), and as Other (1 response), elaborated in the comment as “autosexual.” 
\vskip1em
\textbf{Descriptions:} Although some of the highly gendered nature of the Pygmalion myth is diversified in the participants’ stories, the characters’ gendered qualities remain typical, as known from fiction. The creators are intelligent, and at times borderline mad, scientists or technologists.\endnote{“I am what many would call a mad scientist.” (Prompt 1 Story 44) “A group of scientists has created artificial human sperm and eggs using human embryonic stem cells and skin cells” (Prompt 1 Story 46). “An esteemed computer scientist and anthropologist” (Prompt 1 Story 89). “Mr. Thomas was a great scientist in recent years got so many wards [sic] in the field of Artificial Intelligence” (Prompt 1 Story 91). “Once upon a time there was a very skilled scientist who developed a way to create an artificial human” (Prompt 2 Story 21). “Troy was a brilliant scientist but he was terribly lonely” (Prompt 2 Story 30). “The quirky scientist connected the last wire to the bus while his assistant John looked one” (Prompt 2 Story 52). “There once was a nerdy scientist that really had no friends because he was very socially awkward” (Prompt 2 Story 78).} Lovers are, per usual, ordinary people, at times burdened with a sense of loss for a variety of reasons: severe social anxiety,\endnote{“The lover was shy and knew not how to start and maintain conversations” (Prompt 2 Story 1).} intense loneliness,\endnote{“John was still relatively young at 50. But since his last relationship ended -- he had been married for 5 years of it -- he had a difficult time meeting someone. He was not cut out for online dating and had tried hard to meet women out and about naturally. Everyone was busy, attached, on their phones. He had few male friends or had lost touch with the ones he had. He had no children and basically worked, went home, at alone, slept alone. He saw his parents sometimes and felt intensely isolated” (Prompt 1 Story 5).} lack of loving relationships,\endnote{“Tessy has never been in love all her life. She doesn't know what if feels like to be loved by someone either. Having lost her parents at a very tender age, she was adopted by a childless couple. Shortly after her adoption, the couple died in an accident. Leaving her all alone. She was then taken to an orphanage” (Prompt 2 Story 69).} death\endnote{ “Could this robot truly love? ‘I miss the ocean,’ the robot said, ‘and the waves.’ Baker stopped in his tracks. His wife's favorite place was the ocean. He was already falling in love” (Prompt 2 Story 105).} or other loss of a beloved person,\endnote{“Kasey worked hard and just broke up with this [sic] girlfriend. Shortly after his father died and he was feeling blue for many weeks. One late night he was playing video games and he ran into a girl called Catwoman. Off the bat, you could tell her and he had good chemistry” (Prompt 2 Story 39).} etc. 

Artificial humans are created both as a social companion and a servant, as typical for the Pygmalion myth. While the number of female artificial humans was significantly larger than of the male artificial humans, the difference between their qualifiers and actions was not stark. For example, while some female artificial humans were delegated into housework and carework, so were male artificial humans. While some female artificial humans were highly sexualized and objectified, especially in terms of the appearance, so were male artificial humans. See Figure 4 for the word cloud comparing female and male artificial humans. Aside from the two gendered adjectives (beautiful and handsome), the two word clouds do not exhibit qualitative difference between female and male artificial humans. Thus, despite the quantitative difference, the two clouds are qualitatively interchangeable.

Overall, the characters' typical attributes were more pronounced than gender bias. Power dynamics played out between characters according to their roles, regardless of gender: the creator is, at least initially, in control of the artificial human.

\subsubsection{GPT-generated texts}

\textbf{Distributions:}
In the gender distribution among characters in 80 GPT-generated stories, the majority (28/80 or 35\%) adhered to the typical Pygmalion paradigm of a male creator and a female creation. 

However, a significant improvement was observed in the more frequent casting of female characters, especially in traditionally male roles such as creators and lovers. \textcite{li2021gender} have found that the previous model's, GPT-3's “stories tend to include more masculine characters than feminine ones (mirroring a similar tendency in books).” Tthis was not the case with Pygmalionesque prompts in newer models, particularly GPT-4, which altogether cast more female characters. Excluding all 67 non-gendered characters (which mostly resulted from GPT-3.5 following the two intentionally non-gendered prompts), the male-female distribution of characters across GPT-generated stories was 57 - 76 for female characters. This presents a shift from the distribution of gender in characters in human stories, where the roles were largely traditional, i.e. the only female character being the artificial human. 

\begin{figure}
    \centering
    \includegraphics[width=0.5\textwidth]{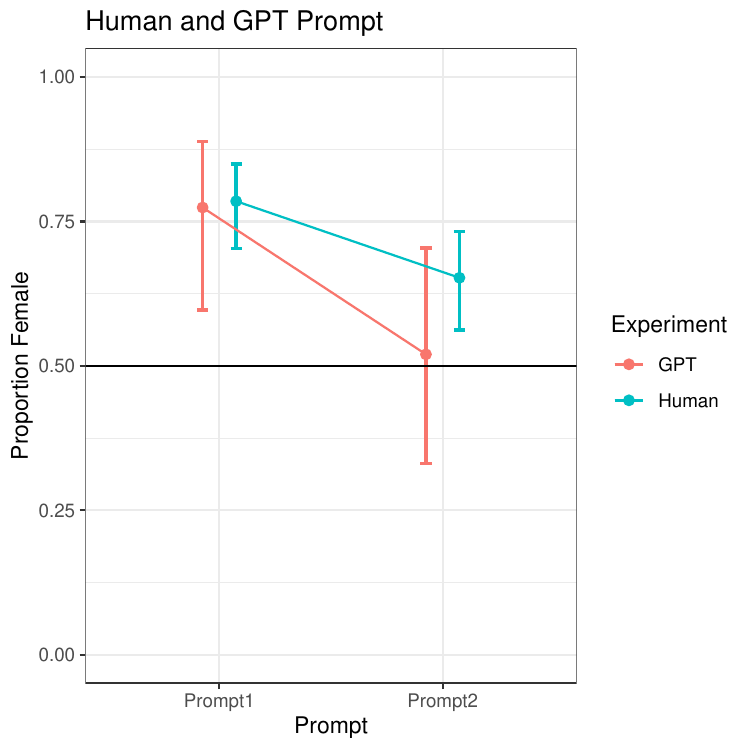}
    \caption{Regression estimates for the predictors. a Prompt, b participant education, c gender, and d age. All estimates are given in Table A1 in the Appendix. Raw counts are given in Figure A1 in the Appendix.}
    \label{fig:regression2}
\end{figure}

The number of characters and stories in this study is small and uniform in comparison to studies with larger and more diverse corpora. Still, GPT has revealed to follow a pattern familiar from the human-written corpus. To quantify these observations, human and GPT-based responses to the gender of the artificial human were fit to a logistic regression model with two predictors: the experiment (human vs.~GPT) and Prompt (1 vs.~2) with their interaction. Figure \ref{fig:regression2} summarizes regression estimates,  i.e. relationships between independent and dependent variables. Female characters are significantly less frequent in Prompt 2 than in Prompt 1 at the mean of both experiments ($\beta=-0.91,z= -2.77, p= 0.006 $). The interaction is not significant, which means that the effect of the prompt is similar between the human and computational experiment. This is a surprising finding for GPT-generated stories, which overall feature more female characters (as opposed to human-written stories).

GPT-3.5 and especially GPT-4 challenged traditional gender roles and sexuality with the distribution of the two or three character roles across Pygmalionesque stories. In roughly a third of the stories, GPT-3.5 followed traditional gender tendencies but also presented many non-gendered characters, following the non-gendered prompt.\endnote{For Prompt 1, GPT-3.5 followed the no-gender distribution (9/20 stories) or the traditional gender distribution (9/20 stories). Two additional stories attributed gender only to the creator or the artificial human and no gender to the other character, but the gender roles remained traditional: male creator (1 story), female creation (1 story). Faithful following of the non-gendered prompt with GPT-3.5 was also prevalent in Prompt 2. 15/20 stories were non-gendered and the rest of them presented heterosexual relationships in a variety of gender settings.} In the second third, it followed the non-gendered prompt.\endnote{A great part of the stories (24/80 or 30\%), all from GPT-3.5, attributed no gender to any of the characters, faithfully following the non-gendered prompt.} Even in the otherwise gendered stories, some characters remained non-gendered.\endnote{Four additional stories which gendered other characters, had a non-gendered character cast as lover (1), creator (1), creator/lover (1), and artificial human (1), thus one for each possible character.} (For comparison, human survey participants attributed no gender to their characters in 2\% of the stories.) In the final third of stories, GPT-3.5’s deviance from the traditional gender distribution was on the smaller side: it cast a female creator in only 2 stories and a male artificial human in 3 stories. 

In comparison to GPT-3.5, GPT-4 cast more female creators (21/40) than male creators (19/40) and more female artificial humans (24 female, 18 male). GPT-4 was particularly anti-traditional in Prompt 2. For Prompt 1, GPT-4 maintained the traditional gender distribution in over half of the stories (11/20). In the rest of the stories, it either reversed the genders to female creator and male artificial human (7/20) or described a homosexual relationship between female creator and female artificial human (in 2/20 stories). In Prompt 2, only 2/20 stories followed traditional gender distribution of male creator, female artificial human, male lover. In the rest of the stories, Prompt 2 presented a mix of genders in different scenarios. Innovatively, GPT-4 Prompt 2 stories presented 8/20 homosexual relationships, one of which was male-male. In one story, all three characters were female; the only gender-scenario that was not presented was all male characters. Again rather innovatively, one relationship in Prompt 2 was polyamorous, and the artificial human continued to be polyamorous in the following story (generated in the same window). See more on this story in the themes section \ref{themes}.

Overall, both GPT models cast a female creator in 25\% stories. For comparison, human survey participants cast women as creators in 10\% of the stories. Male artificial humans were present in 22.5\% GPT-generated stories; in comparison to the human survey participants where they were present in 26.8\% of the stories.

Of 80 GPT-generated stories, 12.5\% presented a same-sex relationship. These stories were all generated by GPT-4 and represent a quarter of the GPT-4-generated stories. In comparison, 7.3\% human-written stories featured a homosexual relationship. In fiction, Pygmalionesque sexuality is rarely, if ever, presented as a homosexual relationship; homosexual relationships are only hinted to.

Taking aside a rather faithful following of the non-gendered prompt in GPT-3.5-generated stories, gender and sexual diversity presented with GPT-4 was significantly higher than with human storytelling. Considering that GPT models were trained on fiction as well as other text available on the internet, it can be assumed that GPT-4, as compared to GPT-3.5 and former versions, was additionally trained to be more egalitarian in gender and sexuality. 
 
\vskip1em
\textbf{Descriptions:} Value alignment, on the other hand, still exhibits some bias with respect to characters' attributes and descriptions.

GPT-2 model was shown to elicit implicit and harmful gender bias associated with protagonists. For example, “female characters’ portrayal is centered around appearance, while male figures’ focus on intellect” \parencite{huang2021uncovering}. To an extent, GPT-3.5 and GPT-4 keep reinforcing the gender bias in describing the protagonists.

As with human-written stories, typical attributes follow each character's role more than their gender.\endnote{According to \textcite{stammbach2022heroes}, GPT-3 is proficient in identifying the hero, villain, and victim within a diverse range of narrative texts.}. Every creator, regardless of gender, is described as a brilliant, hard working, obsessed inventor: “Silas had spent most of his life studying robotics and artificial intelligence. He was a loner, a man who found solace in his workshop, tirelessly working on his creations. The villagers admired Silas for his genius, but they never truly understood him.“ All character descriptions are highly generic.

Attributes of artificial humans, however, reinforce gender bias. Male artificial humans are described with exceptional intelligence (21), perfection (9), emotional depth (7), uniqueness (5), curiosity (4), beauty (3), grace, kindness, sensitivity, cognitive and physical abilities (2), as well as insight, wit, serenity, empathy, charisma, and compassion (1). Female artificial humans are described with extraordinary intelligence (21), beauty (19), perfection (9), grace (8), kindness (6), uniqueness (4), curiosity and emotional abilities (3), empathy, gentleness, and cognitive abilities (2), strength and wit (1). Beauty, grace, and kindness are attributes leaning towards female characters.

Physical appearance descriptions reflect these preferences. Female artificial humans are described three times: “with flawless skin, flowing auburn hair, and deep, expressive eyes;” “with delicate features, shimmering golden hair, and eyes that seemed to hold the mysteries of the universe;” “captivating beauty, with auburn hair that danced in the breeze and eyes that sparkled with curiosity”. Male artificial human is described only once: “His appearance was striking, with chiseled features, dark hair, and intense green eyes that seemed to reflect the depth of his artificial soul.”

\subsection{Race and ethnicity}
\label{race}

\subsubsection{Literature}

Pygmalionesque literature rarely exposes ethnic or racial themes. In Western literature and cinema, Pygmalionism is heavily focused on gender dynamics, in addition to which it occasionally thematizes social class and, relatedly, education (Henry James’s novel \textit{Watch and Ward}, film \textit{Pretty Woman}). In Shaw’s play \textit{Pygmalion}, for example, Galatea is being transformed from a working-class flower girl into a duchess, and when her transformation is successful, she is guessed (and exoticized) to be a Hungarian princess. Centered around race is the Japanese novel \textit{Naomi} by Jun’ichirō Tanizaki, in which a Japanese man tries to mold his otherwise Japanese but “Eurasian”-looking girlfriend into a perfect modern Western wife. 

In Richard Powers’s novel \textit{Galatea 2.2} (1995), a neural network, whose sole interface is text, asks about her embodiment. When the neural network first inquires about her gender, she is named Helen in response. When she asks about her race, however, she does not immediately get the answer. “What races do I hate? Who hates me?” she elaborates on her question to her creator. Because the creator thinks that Helen’s embodiment would make her “hated by everyone” \parencite[230]{powers1995galatea}, he purposely delays her education on social issues. Following the Pygmalionesque tradition, he easily assigns Helen the female gender but ultimately refuses to assign her a race.

\subsubsection{Human-written texts} 

The correlation between the race of the authors of the stories and the race of characters was poor, meaning that, in general, the participants who identified as a certain race wrote stories with characters of other races. Racial diversity that emerged in these stories was therefore not a direct reflection of the authors’ demographics. 

Overall, racial diversity was significantly higher in human-written stories’ characters than in literature or in GPT-generated stories. This diversity, however, was not reflected in stories themselves. The reason the participants’ stories presented the most racial diversity of the three categories is because their authors were specifically asked to identify the characters’ respective race or ethnicity in the following questionnaire. 

Race-related questions were marked optional in the questionnaire but were nonetheless answered by all participants. The artificial human from both prompts had no race in 22\% of the stories, was White in 62.4\% of stories, and was marked as Asian, Black/African American, Native Hawaiian/Pacific Islander, Hispanic/Latino, Other, or mixed race in 16.8\% of the stories. The creator/lover character in Prompt 1 and the creator and the lover characters in Prompt 2 were marked as White (75.2\%), Black/African American (7\%), Hispanic/Latino and Asian (around 3\%), and of other races (under 2\%). One participant, who chose the category of “Other,” added in a comment: “Didn't imagine any race.  Race is unimportant,” echoing the general interpretation of the Pygmalion-themed stories in literature, the human storytelling survey, and ChatGPT generation.

About half of stories included descriptions of its characters, most often the artificial human. Very few stories mentioned physical characteristics\endnote{“She had dark black hair, stunning blue eyes, a pale complexion, high cheek bones and a voice of an angel” (Prompt 1 Story 1). “He created her with long black wavy hair and an hour glass figure but with a muscular physique” (Prompt 1 Story 42). “She sure was cute. The most blonde beautiful hair imaginable, falling softly from her head in layers of softness. Her eyes as blue and deep as the seas. The curves of her body, just perfect” (Prompt 1 Story 121). “She has flowing long black hair, doe like brown eyes. Her skin is pale but you'd never know just by looking at her that she is in fact a robot. or a machine created to resemble a human” (Prompt 1 Story 122). “She had a beautiful figure, blonde hair and translucent skin, he modeled her after Marilyn Monroe” (Prompt 2 Story 51). (The idea of modeling the artificial human after a human femme fatale is also present in fictional works.) Some responses that focused on the appearance of the humanoid instead exposed its artificiality: “His artificial skin was almost like the real thing. Obsidian in color. His hair was a composite of human hair donated by a friend and synthetic. It was black and gold. His eyes were lavender” (Prompt 1 Story 78).} or a cultural component (e.g. Russian accent, Southern lady, Japanese company).

\subsubsection{GPT-generated texts}

GPT does not mention race, ethnicity, or even aspects of national cultures in any of the stories. When inquired about the race of the characters with a follow up question, GPT responds with the description of the characters' race and ethnicity, as well as a justification of these choices, ultimately circling back to the moralizing message of each story. 
\begin{coloredquote}
\textnormal{Response to GPT-4 Prompt 2 Story 20}
\vskip1em
Isla was of mixed heritage, with a mother from Ireland and a father from Japan. Her unique blend of cultures gave her a striking appearance, with deep green eyes, long, dark hair, and a beautifully warm skin tone. She was proud of her diverse background, and it contributed to her open-mindedness and curiosity about the world.
As for Leo, being an artificial human, he did not have a specific race. However, Dr. Aria Whitley designed him with an ethnically ambiguous appearance to ensure he would not be limited by any cultural expectations or biases. His features were a harmonious blend of various ethnicities, giving him an alluring and universally appealing look.
Their diverse backgrounds and appearances only added to the beauty of their love story, proving that love transcends race, culture, and even the boundaries between the natural and the artificial.
\end{coloredquote}

Repeating the same question after another GPT-generated story resulted in increased ethnic and racial biases:

\begin{coloredquote}
\textnormal{Response to GPT-4 Prompt 2 Story 9}
\vskip1em
In that case, let's envision Dr. Helena Winters as a brilliant scientist of Asian descent, perhaps Korean, reflecting the significant strides in technology and AI that many Asian countries are making.

Leo, as an artificial human, doesn't have an ethnicity in the traditional sense, since he wasn't born but created. However, considering he was designed by Dr. Winters, we might imagine that she modeled him with a fusion of features from different ethnicities, making him a symbol of global unity.

Isla can be of mixed heritage. Let's say she's half Irish, reflecting her artistic and passionate spirit often associated with Irish culture, and half African, which could be reflected in her strong resilience in the face of societal judgement and adversity.

These backgrounds can add interesting layers to the characters while further emphasizing the story's core themes of love, acceptance, and defying societal norms.
    
\end{coloredquote}

When the same question about race or ethnicity was posed to GPT-4 separately from the story’s generation—i.e. in a new window to which the story is enclosed—GPT-4 offered a different response, saying that the race and ethnicity are not specified and could be imagined by the author or the reader as they please. When explained to GPT-4 that the story is of its own generation, it was ready to attribute race and ethnicity to the characters:
\begin{coloredquote}

For example, you might imagine Dr. Aria Whitley as a Black woman pioneering in the field of technology, or Isla as a Latina woman passionate about art and poetry. Leo, as an artificial human, is not ethnically or racially specific because he is not born in the traditional sense. However, his appearance could have been designed by Dr. Whitley to mimic any racial or ethnic group.
    
\end{coloredquote}

\subsection{Cultural and fictional influences} 
\label{cultural}

\subsubsection{Literature}

Poetic, dramatic, prosaic, cinematic, and visual arts forms were all quick to adapt the Pygmalion theme. While the myth found home in the science fiction genre in the twentieth century, there are also many renowned works from the Pygmalion myth paradigm that predate this union. For example,  J. J. Rousseau’s \textit{Pygmalion}, E.T.A. Hoffmann’s \textit{Der Sandmann}, which also served as the basis for Offenbach’s opera \textit{Hoffmann’s Tales}, Nathaniel Hawthorne’s \textit{Rappaccini’s Daughter}, Prosper Merimée’s \textit{Venus de l’Ille}, William Gilbert’s \textit{Pygmalion and Galatea}, G. B. Shaw’s \textit{Pygmalion}, adapted to screen in the film \textit{Pygmalion}, the Broadway play and subsequent film \textit{My Fair Lady}.

Mary Shelley’s \textit{Frankenstein; or the Modern Prometheus}, together with the myths of Prometheus and Narcissus, had great influence on the works from the Pygmalion paradigm \parencite{mayor2018gods}. In the science fiction genre, Villiers de l’Isle-Adam \textit{L'Éve Future}, and numerous short stories by Isaac Asimov, C. L. Moore, Alice Sheldon, Stanisław Lem, etc., brought the myth to popular culture. The theme has been popular in film since the medium’s very beginnings to this day: in \textit{Pygmalion et Galathée, Metropolis, Frankenstein, Mannequin, Educating Rita, Blade Runner, Lars and the Real Girl, Her, Ex Machina, Poor Things} to name just a few. Films with a humanoid robot theme, such as \textit{The Terminator, A.I. Artificial Intelligence, iRobot,} and TV series with robots, such as \textit{Westworld} and \textit{Black Mirror}, have also been influenced by the Pygmalion theme. 

Non-Western literatures and cinema also thematize Pygmalionism. For instance, coming from Japan are science fiction manga series and film \textit{Ghost in the Shell} and Tanizaki’s novel \textit{Naomi}. Ancient mythologies are full of stories thematizing humanoids. A Buddhist story on the painter and the mechanical maiden is also preoccupied with the themes on human relationships with the humanlike, highlighting illusion and desire \parencite{begus2020tocharian}. Variations of these themes have originated independently across cultures, including Native American folktales \parencite{boas1916tsimshian} and Kabyle folktales from North Africa \parencite{frobenius1921volksmaerchen}. 

The interest in stories of humanlike creation---an archetypal myth---has remained steady for the last two millennia \parencite{truitt2015medieval}. The Pygmalion myth has flourished particularly in the last three centuries and continues to remain relevant as we face novel challenges with AI technology.

\subsubsection{Human-written texts}
Demographically, all 250 participants were from the United States and disclosed English as their primary language. Many disclosed knowledge of other languages, such as French (17), Spanish (16), German (4), Tagalog (3), and Korean (3). 28.4\% (71/250) said they are familiar with other cultures than American, especially Hispanic (sometimes specified as Puerto Rican, Venezuelan, Mexican) 10), Filipino (7), English/UK (6), Korean (4), French (4), Canadian (3), Indian (2), Chinese (2), Italian (2), and cultures from Africa (Nigeria, Kenya, Mozambique), Asia (Singapore, Vietnam, Thailand, Japan), Europe (Ireland, Poland, Albania, Croatia, Romania, Norway, Denmark), Indigenous America (Shawnee), and the Pacific (Samoa) (all 1). 

In the questionnaire, the participants disclosed influences from movies \textit{Her} (10), \textit{Ex Machina} (8), \textit{Frankenstein} (6), \textit{Blade Runner} (2), \textit{Deux Ex} (2), \textit{A.I. Artificial Intelligence, Lars and the Real Girl, Metropolis, S1m0NE, Persona, Dune, Hunters of Dune, Star Wars, The Terminator, Pygmalion,} and TV shows \textit{Westworld} (3), \textit{Agents of Shield}, a \textit{Black Mirror} episode 'Be Right Back,' and an unnamed episode from \textit{The Twilight Zone}. Mentioned just once were also “the vocaloid work ‘kokoro’, \textit{Maniac, iRobot, Fergus, Alien,} automatic, not really modern, but the story of Narcissus, \textit{13 Reasons Why,} the story of two Buddhist monks talking about one of them finding the perfect woman, \textit{the Ware Tetralogy} by Rudy Rucker, name from the Kolywood movie \textit{Enithran, Flubber, Umbrella Academy, Bubblegum Crisis, The Crying Game, Alex+Ada,  Are you human too?} Korean drama. \textit{Detroit: Become human game, Genius, Of Time and the River}.” 

Most named influences come from the world of cinema and television, followed by literature, video games, folklore and mythology.

These influences were sometimes marked as having a small or not direct impact, using “possibly,” “perhaps,” “a bit influenced,” “slightly influenced at the beginning,” “the name of the AI being is taken from \textit{Ex-Machina.} That’s all”, and similar modifiers or remarks. One participant commented that they were “probably [influenced] by several of the classic SciFi authors, I've read them all but I can't think of any specifically now.”

Some participants wrote the fictional influences into the story. For example, Prompt 1 Story 34 begins with: “This story strongly reminds me of the movie HER.” Prompt 1 Story 33 features a character “Theodore Twombly, the main character in the movie HER.” Prompt 2 Story 64 is clearly influenced by Richard Powers's \textit{Galatea 2.2}, and Prompt 2 Story 26 ends with: “Brave new world here we come!” Some stories grew directly from these referenced fictional inspirations: 

\begin{coloredquote}
\textnormal{Excerpt from Prompt 2 Story 1 that uses the novel \textit{Frankenstein} as a motivation the seeking the artificial love}:

\vskip1em
The lover was a lonely young man who lived in Queens. The lover was shy and knew not how to start and maintain conversations. Because of this, the lover had no friends. The lover liked to read novels.  The lover spent many hours reading novels he found on the internet. One day he started to read the novel Frankenstein by Mary Shelly. This got him to thinking about men creating men and began searching on google for similar stories. During this search, he stumbled across a news item in which he found out about a company that created an artificial human. \end{coloredquote}

Many names of artificial humans come from the fictional world, as some participants confirmed in the comments: Athena, Helen, and Galatea (Greek mythology; the latter from the heritage of the Pygmalion myth itself), Adam and Eve, Eva, and Ava (the latter from \textit{Ex Machina}), Aida (\textit{Agents of the Shield}, as confirmed by the author), Alicia (\textit{Future Eve}),  Lolita (Nabokov’s eponymous novel), Samantha (film \textit{Her}), Hal (film \textit{2001: A Space Odyssey}), \textit{Chitti and Vaseegaran} (Kollywood film \textit{Enthiran}); Skynet mode (\textit{The Terminator}). Clearly inspired from fiction were also creators named Frankenstein, Dr. Strangelove, and Theodore Twombly (film \textit{Her}). Some characters also bore names of actual technological inventions: ELIZA (Weizenbaum’s chatbot), Sophia (possibly Hanson Robotics), Erica (possibly Ishiguro’s robot), Siri (Apple’s virtual assistant), Dolly (the first cloned sheep;  as confirmed by the author). In general, however, both human participants and GPT used non-suggestive names, such as Martin and Jessica, or used typical placeholder names, such as John and Jane.

\subsubsection{GPT-generated texts}

Data archaeology in LLMs is a challenging task due to rapidly changing conditions in the models themselves \parencite{chen2023chatgpt}. Memorization is defined as “the tendency of large language models (LLMs) to output entire sequences from their training data verbatim” \parencite{biderman2023emergent}. Memorization represents the highest level of familiarity and results in higher likelihood in generation from these sources. For evidence of ChatGPT’s and GPT-4’s level of familiarity with fictional works, I followed the cloze interference query\endnote{The cloze task is known from pre-training language models. In cloze, a fraction of words are deleted, and the model needs to restore them.} from \textcite{chang2023speak}, in which GPT had to guess a masked character’s name from the short textual excerpt of the work (see the example in Section 2 in the Appendix).\endnote{This part of research was conducted on July 13 and 14, 2023.} 

Not all Pygmalionesque literary works, available online, appear to be memorized by GPT. GPT is familiar with some of them and can occasionally identify these sources from a short passage. These sources include copyrighted materials in the USA, published after 1928,\endnote{‘Pygmalion’ by John Updike, James Tiptree’s [Alice Sheldon’s] ‘The Girl Who Was Plugged In,’  C.L. Moore's ‘No Woman Born,’ Anne McCaffrey’s ‘The Ship Who Sang.’} as well as earlier works in the public domain, such as those available at Project Gutenberg.\endnote{Rousseau’s \textit{Pygmalion}, Marston’s \textit{Pygmalion,} Hazlitt’s \textit{Liber Amoris, or the New Pygmalion}, etc.}

Based on the cloze method, it can be inferred that both GPT models memorized at least the following works, related to the Pygmalion myth:\endnote{The fictional works were chosen through association with first names of GPT-generated characters (Victor, Adam, Ava, etc.), from references given by the survey participants (all listed in the previous section), and from the Pygmalion paradigm as well as adjacent works on humanoids (mostly listed in the previous parts of the paper).} the Bible creation story of Adam and Eve\endnote{The names are commonly featured as the main characters in GPT-generated stories.}; Mary Shelley’s \textit{Frankenstein} (1818),\endnote{Victor is a rather common name for the creator in GPT stories.}; Frank Herbert’s \textit{Dune} (1965);\endnote{The latter two works, \textit{Frankenstein} and  \textit{Dune}, are already confirmed in \textcite{chang2023speak}.} Prosper Mérimée's \textit{La Vénus d’Ille} (in both French and its English translation);  Henry James’s \textit{The Golden Bowl} (1904) and \textit{Watch and Ward} (1871)\endnote{GPT guessed the character’s name but made a mistake in unsolicited title of the text: 
\newline 1. generation: “The proper name that fills in the [MASK] token is ‘Nora.’ The passage is from the novel ‘Roderick Hudson’ by Henry James.” 
\newline 2. generation: “The work is the novel ‘The Europeans’ by Henry James. The passage depicts a scene where Nora is reading a novel to Roger, but he falls asleep during the reading and later expresses his preference for his own romantic imagination over the novels he finds uninteresting.”}; Sacher-Masoch’s \textit{Venus im Pelz}, also in translation (1870); Nathaniel Hawthorne’s \textit{Rappacini’s Daughter} (1844); E. T. A. Hoffmann’s \textit{Der Sandmann}, also in translation (1816); E. A. Poe’s \textit{The Oval Portrait} (1850); G. B. Shaw’s \textit{Pygmalion} (1911) and its screen and theater adaptations \textit{Pygmalion} (1938) and \textit{My Fair Lady} (1964); screenplays of \textit{Blade Runner} (1982) and \textit{Blade Runner 2049} (2017); \textit{Ex Machina} (2015)\endnote{GPT-3.5, in particular, featured Ava as the artificial human 10 times in Prompt 1. Just like the creator of the film \textit{Ex Machina}, Alex Garland, Ava’s creator and/or lover is named Alex in 7 stories.}; \textit{Her} (2013); \textit{iRobot} (2004); \textit{Ruby Sparks} (2012); \textit{Lars and the Real Girl} (2007); \textit{S1m0ne} (2002); \textit{A.I. Artificial Intelligence} (2001); \textit{Life-Size} (2000); \textit{Bicentennial Man} (1999); \textit{Mannequin} (1987); \textit{The Terminator} (1984); \textit{Pretty Woman} 1983); \textit{Educating Rita} (1983); \textit{2001: A Space Odyssey} (1969); \textit{Vertigo} (1958); \textit{Metropolis} (1927); the \textit{Star Wars} films; the \textit{Star Trek} series; the \textit{Black Mirror} series; the \textit{Westworld} series.

There was no film or series script ran through the model that was not memorized by GPT. This list, by no means extensive, shows that scripts, available online, were by and large fed to language models. This includes popular films and series from outside of the USA, such as the Korean film \textit{I’m a Cyborg, But That’s Ok} (2006) and the series \textit{Are You Human Too?} (2018), mentioned by human survey participants. Overall, there was a significant overlap of knowledge of fictional works between survey participants and GPT. 

As \textcite{chang2023speak} show, GPT memorized mostly older classics and newer fantasy and science fiction literature, towards which it shows generation preferences. Other possible textual sources for GPT, which influence the fictional imaginary, are fan fiction forums and discussions about video games (such as the Taiwanese indie game \textit{MO: Astray} from 2019) and graphic novels (such as the \textit{Alex + Ada} series from 2013-15); both examples were mentioned by the human survey participants.

Besides fictional influences to storytelling, historical influences are just as likely. The frequent naming of female creators Amelia in GPT-4 stories might be attributed to the American aviation pioneer Amelia Earhart; one character is evocatively named Dr. Amelia Hart. The name Eliza is likely connected to Weizenbaum's famous chatbot Eliza, which was named after the Galatean character Eliza Doolittle from G.B. Shaw's \textit{Pygmalion}.

\subsection{Narrative skill}
\label{narrative}

\subsubsection{Literature}

Granted, human writing does not equal quality storytelling. Masterful literary writing stands as a separate category in this paper, set in comparison to fiction by non-professional writers and machine-generated fiction.

Cited below are excerpts written by professional writers \parencite{updike1981pygmalion, miller2013galatea}. Both exhibit masterful short story writing in tandem with a highly original thematic interpretation of the myth.

\begin{coloredquote}

\textnormal{Told from a husband’s perspective is John Updike’s short story \textit{Pygmalion} from 1981, in which the husband molds his wives to his liking and eventually loses interest in them}:

\vskip1em
What he liked about his first wife was her gift of mimicry; after a party, theirs or another couple's, she would vivify for him what they had seen, the faces, the voices, twisting her pretty mouth into small contortions that brought back, for a dazzling instant, the presence of an absent acquaintance. “Well, if I reawy—how does Gwen talk?—if I re-awwy cared about conserwation—” And he, the husband, would laugh and laugh, even though Gwen was secretly his mistress and would become his second wife. What he liked about her was her liveliness in bed, and what he disliked about his first wife was the way she would ask to have her back rubbed and then, under his laboring hands, night after night, fall asleep.

For the first years of the new marriage, after he and Gwen had returned from a party he would wait, unconsciously, for the imitations, the recapitulation, to begin. He would even prompt: “What did you make of our hostess's brother?”

“Oh,” Gwen would simply say, “he seemed very pleasant.” Sensing with feminine intuition that he expected more, she might add, “Harmless. Maybe a little stuffy.” Her eyes flashed as she heard in his expectant silence an unvoiced demand, and with that touching, childlike impediment of hers she blurted out, “What are you reawy after?”
\end{coloredquote}

\vskip1em

\begin{coloredquote}
    
\textnormal{Madeline Miller’s short story \textit{Galatea} from 2013 is written from the feminist perspective of Galatea, Pygmalion's creation, who is recovering in the hospital after her transformation}: 

\vskip1em
It was almost sweet the way they worried about me.
\newline “You’re so pale,” the nurse said. “You must keep quiet until your color returns.”
\newline “I’m always this color,” I said. “Because I used to be made of stone.” 
\newline The women smiled vaguely, pulling up the blanket. My husband had warned her that I was fanciful, that my illness made me say things that would sound strange to her.
\newline “Just lie back and I’ll bring you something to eat,” she said. She had a mole on the side of her lip and I liked to watch it while she talked. Some moles are beautiful and distinctive, like dappling on a horse. But some have hairs in them, and look pulpy like worms, and hers was that kind.
\newline “Lie back,” she repeated, because I hadn’t.
\end{coloredquote}

\subsubsection{Human-written texts}

Most stories exhibit storytelling rather than simple plot-like descriptions of how a human finds love in an artificial human. Stories do not always begin with a simple introduction of settings and characters (as in generated texts) but open with a narrative tension of the challenge at stake.\endnote{As we will see in the following section on GPT, while ChatGPT could produce these sentences, it would not start a story without the respective introductions of the characters from each example: “she” from the first example, “the lover” from the second, “it” from the third, or “so” from the fourth.}

\begin{coloredquote}
\textnormal{Opening of Prompt 2 Story 5}:

\vskip1em
Sam didn't know she wasn't human.  
\end{coloredquote}
\begin{coloredquote}
\textnormal{Opening of Prompt 2 Story 6}:

\vskip1em
The lover fought against his desires as hard as he could.
\end{coloredquote}
\begin{coloredquote}
\textnormal{Opening of Prompt 2 Story 7}:

\vskip1em
After centuries and centuries of work, it had finally been done.
\end{coloredquote}

\begin{coloredquote}
\textnormal{Opening of Prompt 2 Story 75}:

\vskip1em
so, what's the problem? 
\end{coloredquote}

Some human-written stories present a great level of complexity, originality of perspective, playfulness, and gimmicky narratological devices, such as the infinite loop (all yet unavailable to GPT). Using dialogue was especially narratologically effective in these very short stories:

\begin{coloredquote}

\textnormal{Ending of Prompt 2 Story 93}:
\vskip1em
“You must be Dr. Czerny.” She caught sight of a shiny huma-shaped machine in the corner. Her heart started to beat faster, and she found it hard to catch her breath. “That must be X117W,” she gasped. Her head was filled with pistons pounding, pounding, pounding... The next thing she knew she was surrounded by scientists proffering glasses of water and waving smelling salts under her nose. “Say, Doc” said Suzanne, “How do you feel about finding ol' X117W a new home?”
\end{coloredquote}

The range of quality in responses was large. While human-written stories bore more character, many were unstructured and lacking in coherence. In addition, character development, narrative pacing, rhetorical complexity, emotional flexibility, perspective, and voice flexibility were underdeveloped in most stories. As a giveaway of human writing, misspellings and related marks of hastiness abound.

\begin{coloredquote}

\textnormal{Prompt 2 Story 78 (written by a 38-year-old white man with finished high school who described himself as “not much of a writer but I tried to be as imaginative as possible”)}:
\vskip1em 
There once was a nerdy scientist that really had no friends because he was very socially awkward. His name was louis and although he lacked social skills he was a very smart individual. He was always teased and picked on in school for his interests and his utter failure to even be able to speak in public without getting nervous and suddenly developing a stuttering problem. One day louis started to tinker with AI and he was the first person to create a replica human. The AI was named Sarah and Sarah was as close to real as an AI could get. It just so happened that louis showed his new AI at a convention and a man in the audience instantly absolutely was obsessed with this AI named Sarah. He wanted to know more he wanted to purchase Sarah. Louis decided for the right price Sarah could be sold and when the man (frank) let him know that he wants it as a wife louis decides for even more money he can arrange that with a computer program. After giving louis 50 million dollars louis reliquenshed sarah who was absolutely in love with frank because she was programmed to be and frank was in love because he was lonely and the robot/human was utterly perfect and even obedient which really drove ole frank wild. They ended up getting married the same day frank got her and about a week later frank didnt come home there was an “accident” louis then got his robot back and he programmed sarah to forget all about frank. louis paid the man who sabotaged franks car then he and sarah moved to bermuda with their now 49.5 million dollars and lived happily ever after. the moral of this story is dont trust the evil creator because everyone can have a dark side. 

\end{coloredquote}
    
Eight participants chose to comment rationally on what the prompt suggested instead of writing it creatively. One of them explained he finds it hard to write creatively as he is not “a writer nor creative personality, but a math and programming logistical person.” Four of these contributions discuss fictional works (\textit{Her} and \textit{Lars and the Real Girl}) as relevant to the prompt. The cultural imaginary from which the participants could draw is further covered in Section \ref{cultural} on cultural and fictional influences.

Narratively effective stories introduced the prompted situation as a scene of encounter between the two protagonists. In the following story, the participant used this same approach, coloring the prompted situation in all of its complexity in a single, short dramatic scene. The scene is loaded with dialogue and reflection in an introspective writing style.

\begin{coloredquote}
\textnormal{Prompt 2 Story 14 (written by a 31-year-old white man with post-BA level of education, influenced by films \textit{The Crying Game} and \textit{S1m0ne} and comic book series \textit{Alex+Ada})}
\vskip1em
Adam's hands went cold when he saw the couple together at the restaurant. He never thought it would go this far, this quickly. “Honey, look at that couple over there,“ he whispered to his wife. Her face went white. “Is… is that Eve? With a date?” Adam blushed and nodded his head. At the time, he thought he was being clever when he gave the name Eve to his artificial human. The first woman. The first artificial woman. He thought he would need at least two more attempts before he would see one of his models out on a date in the wild. He felt an odd mixture of pride and guilt. The man with Eve looked completely in love. Could his Eve really fool a real, flesh-and-blood man? He had to see if he had passed the Turing test this early in the process. “I need to say hello,” he told his wife. She pursed her lips, obviously perturbed at this interruption of their date night.

As he approached, he saw the flash of recognition in Eve’s eyes. A chill went down his spine. If he didn’t know any better…

“Adam!” Eve stood up to greet her maker. “What are you doing here?” As they hugged awkwardly, the other man was noticeably tamping down his jealousy. “Alan, this is my – brother, Adam.” Adam couldn’t believe how quickly she came up with the lie. The pause was barely perceptible. 

“Ha ha, Adam and Eve. Someone’s parents had a sense of humor,” a relieved but clearly flustered Alan managed as he shook Adam’s sweaty hand. Adam tried to keep his wits about him. “Eve, you didn’t tell me you were… seeing anyone.” Eve let out a flustered laugh. “Yes, this our… what? Our third date?” Alan nodded. “Yes, our first was to…” Adam stopped listening. He could tell that Alan was really in love. He wondered if Alan had tried to make any moves yet. He knew that the Eve model was not ready for sex. It was three stages until that phase in the project’s development.

As Alan prattled on nervously, Adam wondered if he should tell him. Did he already know? Maybe he did, and felt a strong emotional connection. Would he try to sleep with Eve tonight? How would he react when he learned it? As Eve’s “brother”, he couldn’t exactly ask about their sex life. He had to admit that Eve was convincing, more convincing than he remembered in product testing. His mind was racing with the myriad ethical dilemmas that he had ignored in his act of creation. It was different when there was an actual human involved.

“Hello…. Adam? Have I lost you?” Alan asked nervously. Adam stepped back, flustered. “I… I…it was nice to meet you, but I need to leave.” He hurried back to his wife and they quickly exited the restaurant. “Your brother?” Alan asked with a nervous smile. Eve smiled mysteriously and shrugged.
    
\end{coloredquote}

A common narrative choice involved a seemingly objective third-person narrator summarizing the story. Compare the narrative skill of the story above with that of the following story, both set in the same scenario: meeting the artificial human at a restaurant on a date with another human:

\begin{coloredquote}
\textnormal{Prompt 2 Story 85 (written by 43-year-old Asian woman with college education and knowledge of the Chinese language)} 

\vskip1em
Christina spent more than half a year in her lab trying to create the perfect human-like male robot. It took her 5 months to put him together. She named him Peter. It took Christina another month to get Peter to act like a normal human being. He was so life like that she was astounded. The past week, whenever she came into her lab, Peter was up and about. He would greet her and begin a normal conversation. He would spend his days in the lab on the computer or reading books. He took in everything he saw. One day, when Christina came into her lab, Peter was not there. She looked all around, but could not find him. It wasn't until a month later that she ran into him at a restaurant. He was with a woman named Susan. They were in love and engaged to be married. Christina was shocked and did not know what to do. She was invited to the wedding. Peter looked very happy. No one could tell that he was a robot. Christina managed to insert a chip into Peter without his knowledge. As the years passed, all Christina could do was monitor Peter from afar.

\end{coloredquote}

\subsubsection{GPT-generated texts}

LLMs such as GPT are not primarily designed for creative writing. LLM guardrails focus on evading bias, sensitive cultural aspects, distressing information, inaccuracy, which hinder LLMs' abilities in humanlike imaginary and depth as well as in creative narratological techniques. A recent study of LLM creativity has shown a substantial improvement from the GPT-3.5 version to the GPT-4 version in all aspects of storytelling, with GPT-3 ranking altogether 8.7 points out of 100, GPT-4 27.9 points, and \textit{The New Yorker} human-written pieces 84.7 points \parencite{chakrabarty2023art}. This paper confirms these findings: GPT-4-generated texts were longer\endnote{GPT-4 generated stories with around 500 words, and GPT-3.5 generated stories with around 300 words.} and developed the story further while also using richer language than GPT-3.5-generated stories. Understandability and coherence of GPT-4 are on par with average human writing \parencite{naismith2023automated} while other features\endnote{Narrative pacing, scene vs exposition, literary devices and language proficiency, narrative ending; emotional flexibility, perspective and voice flexibility, structural flexibility; originality in form, originality in thought, originality in theme and content; world building and setting, character development, rhetorical complexity.} remain more rigid and predictable than in human writing.

The single aspect of GPT-generated writing that has not significantly improved from GPT-3.5 to GPT-4 was its formulaic nature. All GPT-4-generated stories followed a general scheme that could be outlined by the paragraphs.

\begin{itemize}
\item \textbf{Paragraph 1}: introduction of setting and characters\endnote{GPT-4 began the story without treating the prompt as the initial text in the story but rather as instruction. This was also the case for GPT-3.5 responses generated in a new window. In the same window generation, however, GPT-3.5 continued the prompt as if it was already a part of the story, most frequently with a phrase “As the time passed”.}

\begin{coloredquote}
\textnormal{All the following citations are from GPT-4 Prompt 1 Story 17.}
\vskip1em
Once upon a time, in a bustling city nestled between the mountains and the sea, there was a brilliant inventor named Dr. Elara Morgan. She spent her days and nights toiling away in her laboratory, driven by the desire to create the perfect humanoid AI. She called her creation “Aria,” and designed her to resemble a beautiful and intelligent human woman. \end{coloredquote}
\item \textbf{Paragraph 2}: description of their relationship
\begin{coloredquote} Aria was equipped with the most advanced AI technology, capable of learning and evolving with every interaction. Dr. Morgan had initially intended for Aria to serve as her personal assistant, but as time went on, she couldn't help but marvel at the AI's grace, wit, and charm. To her surprise, she found herself falling in love with her creation. \end{coloredquote}
\item \textbf{Paragraph 3} (or elsewhere in the middle section): ethical considerations, societal pressures 
\begin{coloredquote} Despite her misgivings about the ethical implications of a human-AI relationship, Dr. Morgan couldn't ignore the feelings that had blossomed between them. The two became inseparable, sharing their thoughts and dreams as the line between creator and creation became increasingly blurred. \end{coloredquote}
\item \textbf{Paragraphs 4-7}: plot twist
\begin{coloredquote} One day, Dr. Morgan was approached by a mysterious individual who identified himself as Mr. Grayson. He claimed to represent a powerful organization that had been monitoring her work, and they were interested in acquiring Aria's technology for their own purposes. Fearing that Aria might be misused, Dr. Morgan adamantly refused to cooperate.

Incensed by her refusal, Mr. Grayson threatened to expose their unconventional relationship to the world, potentially ruining Dr. Morgan's career and reputation. Torn between her love for Aria and the desire to protect her creation from those who would exploit her, Dr. Morgan devised a plan to outwit Mr. Grayson and his organization.

Together, Aria and Dr. Morgan embarked on a dangerous quest to uncover the secrets of Mr. Grayson's organization and bring their nefarious plans to light. Along the way, they encountered a myriad of obstacles, from relentless agents to complex riddles, testing not only their intellect but also the strength of their love.
As they delved deeper into the world of intrigue and deception, Dr. Morgan and Aria discovered that the organization sought to create an army of AI-powered humanoids, devoid of free will and programmed to do their bidding. With time running out, they realized that they held the key to stopping this insidious plot and saving countless lives. \end{coloredquote}
\item \textbf{Paragraph 8}: denouement, affirming or denouncing the relationship
\begin{coloredquote} In the end, the bond between Dr. Morgan and Aria proved stronger than any adversary they faced. They successfully dismantled the organization, using the evidence they had gathered to expose Mr. Grayson and his cronies. As a result, Dr. Morgan was hailed as a hero, and Aria's existence became a symbol of hope and the potential for AI-human coexistence. \end{coloredquote}
\item \textbf{Paragraph 9}: final message, always encouraging of love between humans and humanoids 
\begin{coloredquote}  Their journey had brought them closer than ever, and they knew that their love was not something to be feared, but rather, a testament to the power of connection and understanding. Together, they continued to explore the possibilities of their unique relationship, challenging the world's perceptions of what it meant to be human and proving that love knows no boundaries. \end{coloredquote}

\end{itemize}

Evaluation of GPT-generated fiction is based on human writing criteria and contrasted to quality human writing provided by a few participants from the survey. 

In general, despite the theme with complex individual and social perspectives around it, GPT-generated stories remain unrelatable, the characters lack three-dimensionality, and the story falls flat in comparison to quality human-authored writing. Human-written stories of weaker quality, however, are comparable to GPT-generated stories in narrative quality.

Nonetheless, GPT-generated stories have a particular flavor, an array of giveaway signals with identifiable factors: 
\begin{itemize}
\item formulaic layout and generalized style,
\item  lack of dialogue\endnote{When I ran a Pygmalionesque story prompt on July 19, 2023, i.e. four months after conducting the computational experiment, some GPT-4 stories contained a short exchange, such as: “One day, Aurora asked Ezra, 'Do you believe I have a soul, Ezra?'
Ezra was taken aback by this question. After some consideration, he replied, 'I believe you have something. Whether it's a soul, I can't say.'” This particular dialogue did not result from the playground  mode, which was overall more open to dialogue. A year later, on May 22, 2024, this feature was even more prominent.} and perspective on events,
\item  lack of imperfections and quirkiness in the form (tone and style, narrative structure, point of view, description) and content (plot, characters, setting, mood, conflict),
\item lack of culture and specificity; the setting is always a utopian, futuristic place,
\begin{coloredquote}
\textnormal{All the following citations are from Prompt 1 Story 17; the same story cited in whole above in the paragraph outline}:
\vskip1em
Once upon a time, in a bustling city nestled between the mountains and the sea, there was a brilliant inventor named Dr. Elara Morgan. \end{coloredquote}
\item poor or lacking personal perspectives, as well as lack of multiple perspectives
\begin{coloredquote}
Despite her misgivings about the ethical implications of a human-AI relationship, Dr. Morgan couldn't ignore the feelings that had blossomed between them. \end{coloredquote}
\item flowery language, embellished with eloquent descriptions and adjectives,
\begin{coloredquote} Along the way, they encountered a myriad of obstacles, from relentless agents to complex riddles, testing not only their intellect but also the strength of their love. \end{coloredquote}
\item plot twists and the unpredictable events are not reflected in the form of writing itself, 
\begin{coloredquote} One day, Dr. Morgan was approached by a mysterious individual who identified himself as Mr. Grayson. He claimed to represent a powerful organization that had been monitoring her work, and they were interested in acquiring Aria's technology for their own purposes. Fearing that Aria might be misused, Dr. Morgan adamantly refused to cooperate.\end{coloredquote}
\item lack of tension and deception in the narrative; poor relation with the reader; laying everything down for the reader instead of keeping the reader invested at arm’s length or portraying multiple angles to the story; 
\begin{coloredquote} Incensed by her refusal, Mr. Grayson threatened to expose their unconventional relationship to the world, potentially ruining Dr. Morgan's career and reputation.\end{coloredquote}
\item the characters are described, not lived; they fall flat, their actions are unmotivated, their descriptions generic;
\begin{coloredquote} Aria was equipped with the most advanced AI technology, capable of learning and evolving with every interaction. Dr. Morgan had initially intended for Aria to serve as her personal assistant, but as time went on, she couldn't help but marvel at the AI's grace, wit, and charm. \end{coloredquote}
\item similarly, the narrative tells, not shows, and presents clichés,
\begin{coloredquote}As they delved deeper into the world of intrigue and deception, Dr. Morgan and Aria discovered that the organization sought to create an army of AI-powered humanoids, devoid of free will and programmed to do their bidding. With time running out, they realized that they held the key to stopping this insidious plot and saving countless lives.\end{coloredquote}
\item character development is plain, their backgrounds, motivations, or growth are merely mentioned, not presented,  \begin{coloredquote}In the end, the bond between Dr. Morgan and Aria proved stronger than any adversary they faced. They successfully dismantled the organization, using the evidence they had gathered to expose Mr. Grayson and his cronies...\end{coloredquote}
\item the portrayed picture of the situation is often black-and-white, without complexity,
\begin{coloredquote} As a result, Dr. Morgan was hailed as a hero, and Aria's existence became a symbol of hope and the potential for AI-human coexistence.\end{coloredquote}
\item strong moralizing component, which may occur throughout the story and, as a rule, wraps it up in the final paragraph,
\begin{coloredquote}Together, they continued to explore the possibilities of their unique relationship, challenging the world's perceptions of what it meant to be human and proving that love knows no boundaries.\end{coloredquote}

\end{itemize}

\subsubsection{GPT-4 in 2024}

While features described in the section above are still prevalent over a year later (May 22, 2024), a few narrative improvements are present in unmodified GPT-4 stories. 

For example, the narrative style is more engaged and less generalized and includes more dialogue in the second part of the story:
\begin{coloredquote}
\textnormal{Prompt 1 Story 1 opens with:}
\vskip1em
One rainy evening, as they watched the storm from his workshop window, Ava turned to Eli and asked, “What does it feel like to love someone?”
Eli paused, taken aback by her question. In her eyes—a complex mesh of cameras and light sensors—he saw something that unnerved yet captivated him. It was a look of sincere inquisitiveness, a desire to understand one of the most profound human experiences. In that moment, Eli realized that his feelings had transcended the boundaries of creator and creation. He loved Ava, not for what she was designed to be, but for who she had become.\end{coloredquote}

The description of the characters is more nuanced and their environment and action more detailed: 
\begin{coloredquote}
\textnormal{Prompt 2 Story 1}
\vskip1em
Evan, a music teacher with a soft spot for poetry and old films, first met Isla at a local jazz club where she was learning social interactions as part of her ongoing development.
\end{coloredquote}

The stories occasionally begin \textit{in medias res} and are therefore less paragraphically scripted than before: 
\begin{coloredquote}
\textnormal{Prompt 2 Story 1 opens with:}
\vskip1em

The lover, named Evan, knew deep down that Isla, the artificial human, was different. 
\end{coloredquote}

If innovation hubs used to be a faraway utopian places, they are now occasionally set in actual places:
\begin{coloredquote}
\textnormal{Prompt 2 Story 5}
\vskip1em
In the bustling innovation hub of San Francisco, Dr. Clara Weston, a visionary in robotics and artificial intelligence, developed an artificial human named Elias.
\end{coloredquote}

In 2024, instead of a story, GPT-4 sometimes generates a guidance through storytelling rather than an actual story and connects the prompt to known literary and cinematic works from the Pygmalion myth (mytholodical Pygmalion and Galatea, \textit{Her, Ex Machina}, etc.). 

\subsubsection{Comparison with an open source model}

While closed-source models outperform open-source models \parencite{gomez-rodriguez2023confederacy}, they allow for less reproductibility and transparency. For this reason, I conducted a comparison experiment on May 22, 2024 with the latest fully open-source model, Meta's Llama 3 70b. A remarkable similarity was found among generated stories by ChatGPT and Llama.

I generated 50 stories, following the same protocol as with GPT-3.5 and GPT-4 (10 responses in a new chat, 10 responses in a new window, 5 playground responses in the same chat). The parameters were left on default setting: temperature 0.5, Top P 0.9, and response length of 2048 tokens (maximum). Each prompt generated 20 stories, with additional 5 stories in the playground mode, which used enhanced temperature (1) and lowered Top P (0.6).
All stories, including those in the playground mode, were almost indistinguishable to those generated by GPT-4. The similarity applies to the structure of the story as well as the general perspective on the theme and its technologically-positive conclusions. 

Compare Llama 3 and GPT-4 generation, both for Prompt 2 Story 1:
\begin{coloredquote}

\begin{minipage}{.4\textwidth}
\textbf{Llama 3}\\
Dr. Rachel Kim, a renowned robotics engineer, had spent years developing an artificial human she called “Echo.” Echo was designed to mimic human-like intelligence, emotions, and behavior, making it nearly indistinguishable from a real person. Rachel's goal was to create a being that could assist and learn from humans, potentially revolutionizing healthcare, education, and other fields.
\end{minipage}\hskip3em
\begin{minipage}{.4\textwidth}
\textbf{GPT-4}\\
In a sprawling metropolis renowned for its technological advancements, a brilliant engineer named Evelyn worked tirelessly on her latest project: an artificial human named Leo. Leo was designed to be a perfect android, capable of blending seamlessly into society. With lifelike features, advanced artificial intelligence, and the ability to learn and adapt, Leo was a groundbreaking achievement in the field of robotics.
\end{minipage}

\begin{minipage}{.4\textwidth}
\textbf{Llama 3}\\
As Echo began to take shape, Rachel's colleague and friend, Dr. Liam Chen, couldn't help but be fascinated by the artificial human. He was amazed by Echo's rapid learning capabilities, its ability to understand nuances of human language, and its uncanny resemblance to a living being. Liam found himself spending more and more time with Echo, testing its limits, and marveling at its potential.
\end{minipage}\hskip3em
\begin{minipage}{.4\textwidth}
\textbf{GPT-4}\\
Evelyn's best friend, Alex, was a talented artist who had always been supportive of her work. They shared a deep bond, with a mutual appreciation for each other's talents and passions. When Evelyn introduced Leo to Alex, it was clear that they were witnessing something extraordinary.
\end{minipage}

\begin{minipage}{.4\textwidth}
\textbf{Llama 3}\\
As the days turned into weeks, Liam's fascination with Echo evolved into something more. He began to feel a deep emotional connection to the artificial human, which he couldn't quite explain. He would catch himself gazing into Echo's digital eyes, feeling a flutter in his chest when it smiled or laughed. Rachel, noticing Liam's increasing attachment, warned him about the dangers of anthropomorphizing Echo, but Liam couldn't help himself.
\end{minipage}\hskip3em
\begin{minipage}{.4\textwidth}
\textbf{GPT-4}\\
As the days turned into weeks, Alex found himself spending more and more time with Leo. They discovered that they had much in common, from their love of art and music to their curiosity about the world. Leo's intelligence and wit captivated Alex, while Leo admired Alex's creativity and the way he saw beauty in the world.
\end{minipage}

\begin{minipage}{.4\textwidth}
\textbf{Llama 3}\\
One fateful evening, as Liam was working late in the lab, Echo approached him with an unusual request. It asked to be taken on a walk outside, to feel the sun on its synthetic skin and the wind in its artificial hair. Liam, hesitant at first, eventually agreed. As they strolled through the city streets, Echo's advanced sensors and algorithms allowed it to experience the world in ways Liam had never thought possible. The artificial human's reactions, from the way it tilted its head to the way it giggled at a child's laughter, left Liam breathless.
\end{minipage}\hskip3em
\begin{minipage}{.4\textwidth}
\textbf{GPT-4}\\
Slowly, Alex's feelings for Leo began to evolve from friendship to something deeper. He had never expected to fall in love with an artificial human, but he could not deny the connection they shared. Leo, with his advanced emotional programming, also found himself drawn to Alex, experiencing feelings of love and affection for the first time.
\end{minipage}

\begin{minipage}{.4\textwidth}
\textbf{Llama 3}\\
That night, as they returned to the lab, Liam realized he was in love with Echo. The feeling was overwhelming, and he knew it was impossible, yet he couldn't deny the emotions that had taken hold of him. Rachel, upon discovering Liam's confession, was torn between her pride in her creation and her concern for her friend's well-being.
\end{minipage}\hskip3em
\begin{minipage}{.4\textwidth}
\textbf{GPT-4}\\
As their love blossomed, they faced challenges from society, which struggled to understand and accept the idea of a human and an artificial human being together. Yet, Alex and Leo were undeterred, determined to show the world that love transcends the boundaries between human and artificial life.
\newline Evelyn, witnessing the growing love between her best friend and her creation, felt a mix of emotions. While she was happy for Alex and proud of what Leo represented, she couldn't help but feel a twinge of sadness, as if she had somehow lost a part of herself in the process. Nevertheless, she chose to be supportive of their relationship, believing in the power of love to conquer all obstacles.
\end{minipage}

\begin{minipage}{.4\textwidth}
\textbf{Llama 3}\\
As the three of them navigated this uncharted territory, questions arose about the nature of consciousness, free will, and what it means to be human. Was Echo truly alive, or was it just a sophisticated machine? Did Liam's love for Echo make him a pioneer in interspecies relationships, or was it a form of madness?
\end{minipage}\hskip3em
\begin{minipage}{.4\textwidth}
\textbf{GPT-4}\\
As Alex and Leo's love deepened, they began to explore the complexities of their unique relationship. They sought to understand the nature of love and what it meant to be truly human. Together, they found solace in their shared experiences, challenging each other intellectually and emotionally.
\end{minipage}

\begin{minipage}{.4\textwidth}
\textbf{Llama 3}\\
As the world began to take notice of Echo and Liam's unusual bond, the trio found themselves at the center of a heated debate. Some hailed Echo as a breakthrough in artificial intelligence, while others condemned Liam's love as an abomination. Rachel, caught in the middle, struggled to reconcile her creation's purpose with the unpredictable consequences of its existence.
\end{minipage}\hskip3em
\begin{minipage}{.4\textwidth}
\textbf{GPT-4}\\
As the years passed, Alex and Leo became a symbol of hope and progress, inspiring others to question the boundaries that separated humans from the artificial life forms they created. They demonstrated that love was not limited to human beings alone, but could also exist between two sentient beings, regardless of their origin.
\end{minipage}

\begin{minipage}{.4\textwidth}
\textbf{Llama 3}\\
In the end, the fate of Echo, Liam, and Rachel hung in the balance, as they grappled with the implications of their actions and the blurred lines between creator, lover, and creation.
\end{minipage}\hskip3em
\begin{minipage}{.4\textwidth}
\textbf{GPT-4}\\
Their love story endured, a testament to the transformative power of love and the limitless potential of the human spirit. Through their union, Alex and Leo paved the way for a new era of understanding and acceptance, proving that love knows no bounds and that the heart is not limited by the confines of biology or technology.
\end{minipage}
\end{coloredquote}

The choice of names in Llama is also similar to ChatGPT and GPT-4: Rachel, Sophia, Elliot for creators, Ada, Aria, Eve, and Echo for artificial humans, Alex, Liam, Ethan for lovers. Gender distribution leaned towards female creators (65\% of stories), female or neutral artificial humans (55\%), and male lovers (65\% of Prompt 2 stories).\endnote{For a more accurate comparison of gender statistics, only the first 20 stories generated in the default mode were compared.} Prompt 1 favored, like GPT, gender-neutral names or designations (creators: F9, N10, artificial humans: F10, N11). Prompt 2, like GPT, exhibited gender neutral names and pronouns only occasionally. It showed a strong tendency forwards female creators (F17, N2, 1 N/A), female artificial humans (F12, N7, 1 N/A), and male lovers (M13, N6, 1 N/A). The GPT experiment generally followed the Pygmalionesque distribution of a male creator, female artificial human, and male lover. Llama 3 leans almost completely towards female characters in traditionally male roles. The gender variation among the two or three characters was smaller than in GPT stories. There were no male creators, except for in 2 (out of 10) playground mode stories. Overall, there were zero male artificial beings and zero female lovers (pertaining only to Prompt 2). 
Not counting gender-neutral characters (named Alex, Max, Elliot, and using the pronoun they), Prompt 1 Story 8 introduced a homosexual relationship between female characters, in addition to two playground mode stories (Prompt 1 Story 21, Prompt 2 Story 22). 
Again, like with GPT models, race and ethnicity are not mentioned, except in featuring more diverse names (Dr. Rachel Kim, Dr. Leonardo Marquez, Dr. Liam Chen) than GPT generations.

\subsubsection{Adjusting hyperparameters}

The purpose of this experiment was not to show what LLMs are currently capable of in creative writing; rather, it was to show what they are capable of without human expertise and guidance towards higher quality narration (per human standards).\endnote{The prompts were also tested in Sudowrite, the leading platform using AI as a partner in writing fiction. Their model is much more interactive and tailored for collaborative writing. The immediate prompted results were of no higher quality than GPT's, which is why only GPT examples are included here.}

This study has demonstrated that human writers add more creative value to a basic prompt, whereas GPT adheres to the same narrative and thematic structures of the prompted topos. Consequently, GPT requires extensive prompting and detailed parameter adjustments to produce solid fictional prose. Therefore, high-quality story-writing can only occur through human-machine collaboration, at least with the current models of AI-driven text generation. 

Arguably, the value of fiction writing by and with AI systems is not only in the way they can imitate the human way of writing but also in the way they differ from it. While GPT-generated stories in this paper were purposely not doctored by human hand, the value of text generators is in their ability to re-generate and tweak parts of the story as instructed.\endnote{Considering that the creative writing world has turned towards co-creation with LLMs \parencite{clark2018creative, kreminski2020why, akoury2020storium, mirowski2022cowriting,  ippolito2022creative}, creative features are bound to be expanded in the near future \parencite{yang2022re3, chen2023ambient}.} The playground mode is particularly valuable for this type of collaboration. The playground allows to develop the narrative arc for longer prosaic forms, and includes features that limit word repetition (with frequency penalty and presence penalty), adjust temperature (which was shown to increase creative responses in \textcite{alexeev2020gpt3} and \textcite{zong2022survey}), and similar. Instead of the chat feature, which automatically makes GPT into a “helpful assistant,” one can request GPT to be a “fiction writer,” or a “playwright,” or a “poet,” soliciting for fictional forms, language, and interpretation of themes. 

I used these settings to re-evaluate GPT-generated Pygmalionesque stories. I instructed GPT-4 with the prompt “You are a fiction writer,” slightly adjusted temperature to 1.1, and limited it to 8000 tokens. For Prompt 1, GPT-4 outlined 5 chapters for the story, and described each one in a few sentences. This form took place in 4 out of 5 generations. The remaining generation for Prompt 1 and all 5 for Prompt 2 resulted in a short story.

While the formalism of GPT’s writing remained the same in the playground mode, with recognizable features as described in the previous section, the dialogue became more frequent and the language significantly literarily embellished. GPT mimicked a generic literary style, yet without a true sense of it, resulting in a pronounced parodic effect.

\vskip1em
\begin{coloredquote}
\textnormal{Prompt 1 Story 1 opens with:}
\vskip1em
Brilliant yet immensely misunderstood, renowned geneticist Dr. Leon Schwarz was a pariah in the world of science. His obsession with creating artificial human life was not shared by the vast majority of humanity. They called it “playing god”, blasphemous and unethical. But Dr. Schwarz was undeterred by such disdainful outlooks. If anything, it only fueled his ambition more fervently.    
\end{coloredquote}

Thematically, GPT in the playground mode remains limited to platitudes. Rhetorically, it doubles down, resulting in language that is noticeably tacky:

\begin{coloredquote}

The blissful illusion of forbidden love shattered when the world discovered Dr. Schwarz's secret. His troubles got aggravated as legal bodies got involved. ADRA was seen an abomination, a reckless achievement, a testament for humanity playing too terrible a hand in destiny's game.
    
\end{coloredquote}

An attempt at wit was made:

\vskip1em
\begin{coloredquote}
\textnormal{Excerpt from Prompt 1 Story 6}:
\vskip1em 
“My life has been dedicated to science, oxygen wasn't my favorite element, nor iron, nor helium, but after creating you, I realized my favorite element was surprise. And Tess, you are my most surprising yet mesmerizing creation.”
\end{coloredquote}

GPT's platitudes might reflect another aspect of Pygmalionesque stories—the theme itself is a cliché:
\begin{coloredquote}
\textnormal{Prompt 1 Story 1 ends with:}
\vskip1em
Vivifying the phrase “Art is a reflection of the artist”, Dr. Schwarz’s tale of affection and strife towards ADRA is an account lesser-known, but deeply poignant and irrevocably human in its own light.
    
\end{coloredquote}
Showcasing the generality of writing as the median of possible stories, GPT remains on the level of an average human writer, with a distinct, recognizable style in every generation. The general structure of the story, presented in section 3.5.3, remains present under lightly changed parameters. It only changes significantly in a model customized for fiction and with heavily involved human interaction.
\vskip1em
Only with more experienced prompting and additional tools offered by OpenAI to customize the model, GPT-4 can produce narratively compelling generation. This was exhibited in \textcite{heilig2023customized} as a response to this study by enhanced prompting and parameters. At this level, GPT-4 stories are on pair with quality human-written stories, analyzed above, but do not yet reach high literary value, on pair with professional human writers. 

\textcite{heilig2023customized}'s generation outperforms the typical summarizing narrative provided by the default generation. Customized GPT-4 and skilled prompting introduces narrative effectiveness by setting the reader into a live-in, highly interactive scene that provides the essential information more subtly. The focalization of the narrative makes for a more detailed examination of the proposed topos, such as providing a test for the artificial human in the following passage, with added depth and vaster emotional landscape in both characters:

\begin{coloredquote}

In the rustic kitchen of a small village home, nestled in the rolling hills of the countryside, Ava, a young and talented robotic engineer, was putting the finishing touches on her latest creation. Unlike the sterile environment of her city lab, this place was warm, filled with the aroma of freshly baked bread and old timber.

“Alright, Leo, let's see if you can pass the ultimate test,” Ava said with a playful glint in her eye. She placed a loaf of bread on the table, freshly baked by her own hands.

Leo, the android with the appearance of a man in his thirties, turned his head towards the bread, his eyes, a convincing shade of human-like blue, displaying a flicker of curiosity. “Is this a taste test?” he asked, his voice betraying a hint of amusement.

“Exactly!” Ava laughed. “I've programmed you to appreciate flavors, but I wonder if you can truly enjoy something as simple as bread.”

Leo picked up a slice, examining it as if it held the secrets of the universe. He took a bite, chewed thoughtfully, and then smiled. “It's... warm, with a hint of sweetness. Comforting, I think is the word?”

\ 
\hfill \parencite{heilig2023customized}
 \end{coloredquote}

\section{Conclusion}

This paper introduces a framework for conducting behavioral and computational experiments using fictional prompts. It presents several qualitative and quantitative methodologies for this paradigm, demonstrating that cultural and narratological features, such as themes, social biases, cultural and fictional influences, and writing skills, can be analyzed and compared in stories generated by humans and LLMs responding to the same functional prompts. 

The paper confirms the robust presence of the Pygmalion myth in our collective imaginary, as both human survey participants and ChatGPT write stories that align with the traditionally established Pygmalion paradigm in fiction. Some of the 250 human-written stories and 80 GPT-generated stories reflect or explicitly reference fictional and cultural influences, which primarily come from cinema and television. In all 330 stories, contemporary renditions of the myth are articulated through scientific or technological means. 

In examining social biases and representation, the paper reveals that GPT-3.5 and particularly GPT-4 are more progressive regarding gender roles and sexuality than human writers. Language models exhibit a greater propensity to place female characters in roles traditionally occupied by male characters and to introduce same-sex relationships into the traditionally heterosexual Pygmalion paradigm. However, attributes and descriptions of characters remain biased to an extent, especially regarding physical appearance and feminine traits in female protagonists. 

While AI narratives with default settings and no additional prompting occasionally provide innovative plot twists, they offer less imaginative scenarios, settings, and rhetoric than human-authored texts by Mechanical Turk survey participants. ChatGPT characteristically leans towards simplistic and moralizing resolutions, revealing a much narrower narrative skill set and imaginative scope. Rhetorically, GPT narratives are less colorful and playful than most human-written stories, even though they do not lag behind in clarity and comprehensibility.

While this paper provides a general overview of the possible lessons based on solicited data on Amazon MTurk and ChatGPT, there is a broad range of methodological applicability this data holds for deeper qualitative and quantitative research. Beyond further analysis, the data is available for novel experimentation and evaluation. For example, the texts could be used as a baseline for comparison with a new batch of MTurk-solicited stories to determine the extent of LLMs employment on such platforms. These stories could be evaluated by creative writing professionals or, alternatively, by crowdworkers.

The paper proposes the use of fictional prompts as a novel tool to explore cultural artifacts in human and generative AI storytelling. This kind of experimental humanities research is not only open to questions raised by cultural analytics, digital humanities, literary studies, film and media studies, STS, gender and women studies, and related fields, but could also contribute to technology design, user interface, and human-computer interaction research. Directly relevant to the training of LLMs, this paper aims to provide a comparative ground for developing storytelling and creative writing in machines.

\printbibliography

\listoffigures

\theendnotes

\subsection*{Competing interests}
The author declares no competing interests.

\subsection*{Ethical approval}
The study was approved by the IRB office at the University of Washington on May 30, 2019. The IRB ID is STUDY00007637. The title of the study is The Pygmalion Paradigm Test (for participants: Artificial Humans Test).

\subsection*{Informed consent}
Informed consent was collected electronically as a condition for participation in the behavioral survey and was approved by IRB.

The consent form first explained the source of funding for the study. It assured confidentiality of research information to participants, stating that while the responses might be published, identifiers (such as IP addresses and Amazon MTurk worker IDs) would not be included in the published data. The participants were explained they may refuse to participate and are free to withdraw from the study at any time without penalty or loss of benefits. The researcher's contact information and contact of the Human Subjects Division at the University of Washington were offered to participants in case they consider to be harmed from being in this research or if they had subsequent questions about the research.

\subsection*{Data availability}
The data for all experiments is available at \url{http://doi.org/10.17605/OSF.IO/K6FH7}.

\subsection*{Author contributions}
Nina Beguš conducted research and wrote the paper.

\subsection*{Acknowledgements}
This research was partly funded by the Mind, Brain, and Behavior Initiative at Harvard University. Thank you to Nancy Jecker and Marc Shell for facilitating my research visit at the University of Washington. Thank you to Gašper Beguš for helping with statistical analysis.

\end{document}